\documentclass[letterpaper]{article}    % DO NOT CHANGE THIS
\usepackage{aaai2026}                    % DO NOT CHANGE THIS  %%%%%%%%不使用匿名

\usepackage{times}                      % DO NOT CHANGE THIS
\usepackage{helvet}                     % DO NOT CHANGE THIS
\usepackage{courier}                    % DO NOT CHANGE THIS
\usepackage[hyphens]{url}               % DO NOT CHANGE THIS
\usepackage{graphicx}                   % DO NOT CHANGE THIS
\usepackage{natbib}     % DO NOT CHANGE THIS AND DO NOT ADD ANY OPTIONS TO IT
\usepackage{caption}    % DO NOT CHANGE THIS AND DO NOT ADD ANY OPTIONS TO IT
\usepackage{amsmath,amsfonts,amssymb,amsthm} % 基础数学包
\usepackage{booktabs}
\usepackage{threeparttable}
\usepackage{multirow}
\usepackage{bm} % 加粗符号

\usepackage{color}
\usepackage{colortbl}
\usepackage{xcolor}
\usepackage{tcolorbox}

\makeatletter
\newif\if@restonecol
\makeatother

\usepackage[linesnumbered,ruled,vlined]{algorithm2e}
\usepackage{algpseudocode}

\SetKwComment{Comment}{$\triangleright$\ }{}
\SetKwRepeat{Do}{do}{while}

\DeclareMathOperator*{\argmax}{arg\,max}

\renewcommand{\paragraph}[1]{\smallskip\noindent\textbf{#1.}}
\renewcommand{\subparagraph}[1]{\smallskip\noindent\textbf{\underline{#1.}}}

\newtcolorbox{graybox}{
  colback=gray!10,
  colframe=gray!50,
  boxrule=1pt,
  arc=3pt,
  left=5pt,
  right=5pt,
  top=2pt,
  bottom=0pt
}

\title{Partially Shared Concept Bottleneck Models}

\author{
    Delong~Zhao\textsuperscript{\rm 1,}\thanks{Equal contribution.},\;
    Qiang~Huang\textsuperscript{\rm 1,}\footnotemark[1],\;
    Di~Yan\textsuperscript{\rm 1},\;
    Yiqun~Sun\textsuperscript{\rm 2},\;
    Jun~Yu\textsuperscript{\rm 1,3,}\thanks{Corresponding author.}
}
\affiliations{
    \textsuperscript{\rm 1}Harbin Institute of Technology (Shenzhen), Shenzhen, China \\
    \textsuperscript{\rm 2}National University of Singapore, Singapore,\\
    \textsuperscript{\rm 3}Peng Cheng Laboratory, Shenzhen, China \\
    \{zhaodelong,yandi\}@stu.hit.edu.cn, \{huangqiang,yujun\}@hit.edu.cn, sunyq@comp.nus.edu.sg
}

\begin{document}
\maketitle

\begin{abstract}
%%% background
Concept Bottleneck Models (CBMs) enhance interpretability by introducing a layer of human-understandable concepts between inputs and predictions.
%%% challenges
While recent methods automate concept generation using Large Language Models (LLMs) and Vision-Language Models (VLMs), they still face three fundamental challenges: poor visual grounding, concept redundancy, and the absence of principled metrics to balance predictive accuracy and concept compactness.
%%% methods
We introduce \textbf{PS-CBM}, a \textbf{P}artially \textbf{S}hared \textbf{CBM} framework that addresses these limitations through three core components:
(1) a multimodal concept generator that integrates LLM-derived semantics with exemplar-based visual cues;
(2) a Partially Shared Concept Strategy that merges concepts based on activation patterns to balance specificity and compactness; and 
(3) Concept-Efficient Accuracy (CEA), a post-hoc metric that jointly captures both predictive accuracy and concept compactness.
%%% experiments
Extensive experiments on eleven diverse datasets show that PS-CBM consistently outperforms state-of-the-art CBMs, improving classification accuracy by 1.0\%--7.4\% and CEA by 2.0\%--9.5\%, while requiring significantly fewer concepts. 
These results underscore PS-CBM's effectiveness in achieving both high accuracy and strong interpretability.
\end{abstract}

\noindent\textbf{Code —} \url{https://github.com/7494zdl/PS-CBM}

% \begin{links}
%     \link{Code}{https://anonymous.4open.science/r/PS-CBM}
%     % \link{Datasets}{https://aaai.org/example/datasets}
%     % \link{Extended version}{https://aaai.org/example/extended-version}
% \end{links}

% While recent methods automate concept generation using Large Language Models (LLMs) and Vision-Language Models (VLMs), they still face three fundamental challenges: poor visual grounding, concept redundancy, and the absence of principled metrics to balance predictive accuracy and concept compactness. We introduce \textbf{PS-CBM}, a \textbf{P}artially \textbf{S}hared \textbf{CBM} framework that addresses these limitations through three core components: (1) a multimodal concept generator that integrates LLM-derived semantics with exemplar-based visual cues; (2) a Partially Shared Concept Strategy that merges concepts based on activation patterns to balance specificity and compactness; and (3) Concept-Efficient Accuracy (CEA), a post-hoc metric that jointly captures both predictive accuracy and concept compactness. Extensive experiments on eleven diverse datasets show that PS-CBM consistently outperforms state-of-the-art CBMs, improving classification accuracy by 1.0\%--7.4\% and CEA by 2.0\%--9.5\%, while requiring significantly fewer concepts. These results underscore PS-CBM's effectiveness in achieving both high accuracy and strong interpretability.
\section{Introduction}
\label{sect:intro}

%%% applications
Deep neural networks have achieved remarkable success across a wide range of domains, including computer vision, natural language processing, and speech recognition.
Yet, their opaque decision-making process poses a critical barrier to deployment in high-stakes domains such as healthcare and autonomous driving~\citep{chauhan2023interactive}. 
%%% definition
A promising direction for enhancing interpretability is the Concept Bottleneck Model (CBM) \citep{koh2020concept}, which inserts an intermediate layer of human-understandable concepts between inputs and outputs.
% %%% more support for its effectiveness
% By forcing predictions to pass through these interpretable concepts, CBMs enable a modular and transparent decision-making process, facilitating debugging, auditing, and user trust~\citep{wong2021leveraging, huang2024concept}.

\begin{figure}[t]
  \centering
  \includegraphics[width=0.96\columnwidth]{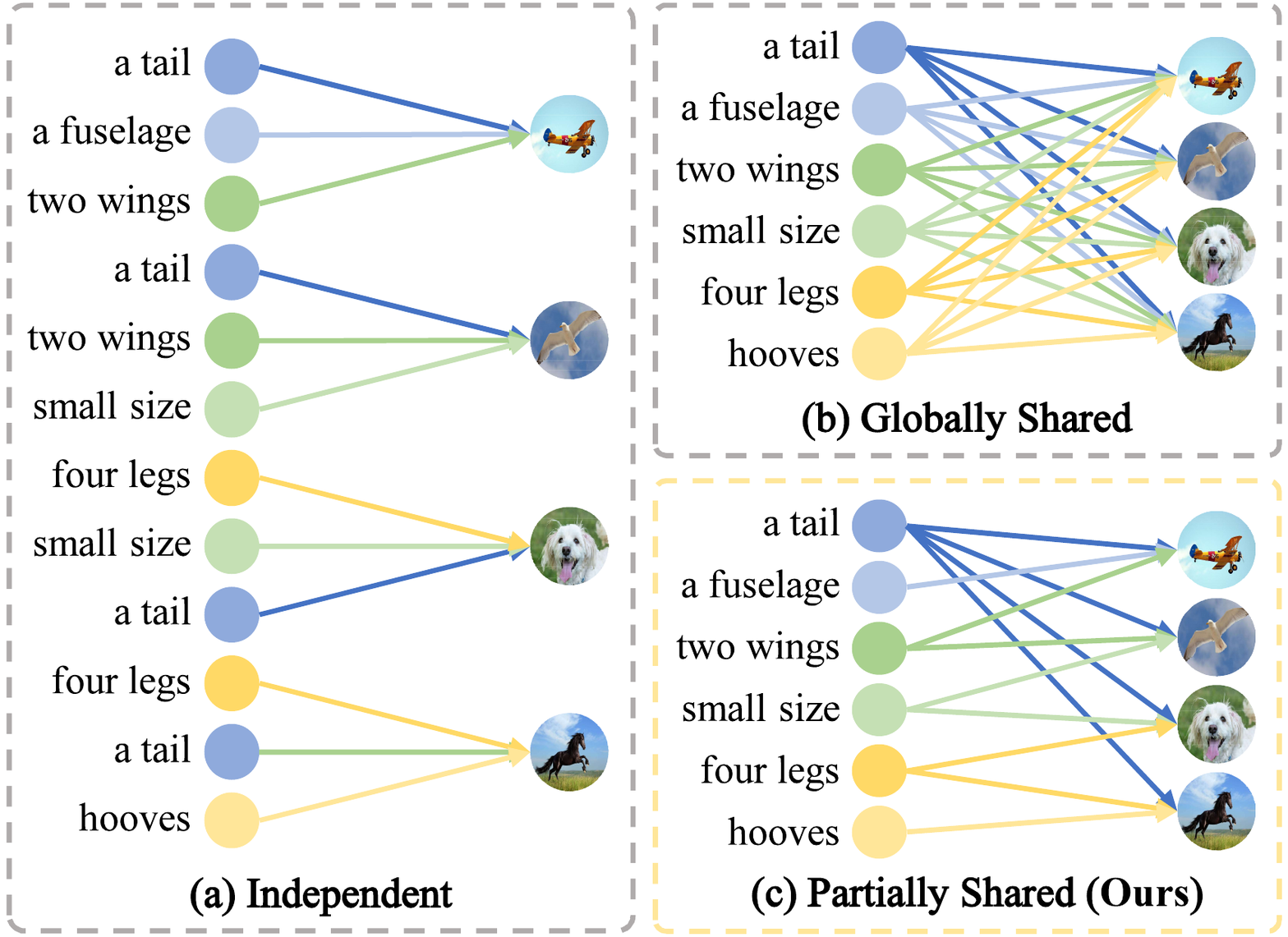}
  \caption{Illustration of different concept sharing strategies: (a) Independent, where redundant concepts exist across classes; (b) Globally Shared, where predictions are affected by irrelevant concepts; and (c) Partially Shared (Ours), which reduces the number of concepts while avoiding interference from irrelevant ones.}
  \label{fig:teaser}
\end{figure}

\begin{table}[t]
\centering
% \small
\setlength\tabcolsep{2pt}
\renewcommand{\arraystretch}{1.2}
\resizebox{0.99\columnwidth}{!}{
\begin{tabular}{lccc}
  \toprule
  \multirow{2}{*}{\textbf{Method}} 
  & \textbf{Semantic-Visual} & \textbf{Low Concept} & \textbf{Principled} \\
  & \textbf{Grounding} & \textbf{Redundancy} & \textbf{Metrics} \\
  \midrule
  \textbf{LaBo} \citep{yang2023language} & $\times$ & $\times$ & $\triangle$ \\
  \textbf{LF-CBM} \citep{oikarinen2023label} & $\times$ & $\times$ & $\triangle$ \\
  \textbf{LM4CV} \citep{yan2023learning} & $\triangle$ & $\checkmark$ & $\times$ \\
  \textbf{DN-CBM} \citep{rao2024discover} & $\checkmark$ & $\times$ & $\times$ \\
  \textbf{Res-CBM} \citep{shang2024incremental}& $\times$ & $\checkmark$ & $\checkmark$ \\
  \textbf{VLG-CBM} \citep{srivastava2024vlg} & $\triangle$ & $\triangle$ & $\triangle$ \\
  \textbf{V2C-CBM} \citep{he2025v2c} & $\checkmark$ & $\times$ & $\triangle$ \\
  \textbf{DCBM} \citep{prasse2025dcbm} & $\checkmark$ & $\triangle$ & $\times$ \\
  \midrule
  \textbf{PS-CBM} & $\checkmark$ & $\checkmark$ & $\checkmark$ \\
  \bottomrule
\end{tabular}}
\caption{Comparison of representative CBMs on three key limitations: Poor Visual Grounding, Concept Redundancy, and Inadequate Metrics. Symbols indicate well ($\checkmark$), partial ($\triangle$), or no ($\times$) improvement.} % Symbols indicate that the challenge is well addressed ($\checkmark$), moderately addressed ($\triangle$), or not addressed ($\times$).}
\label{tab:comparison_single}
\end{table} % vspace
%Formatting-Instructions-LaTeX-2026：There are a number of packages, commands, scripts, and macros that are incompatable with aaai2026.sty. The common ones are listed in tables \ref{table1} and \ref{table2}. Generally, if a command, package, script, or macro alters floats, margins, fonts, sizing, linespacing, or the presentation of the references and citations, it is unacceptable. Note that negative vskip and vspace may not be used except in certain rare occurances, and may never be used around tables, figures, captions, sections, subsections, subsubsections, or references.

%%% prior work before LLMs
While concept-based modeling improves transparency, most early CBMs rely on manually annotated concepts curated by domain experts, which is labor-intensive and difficult to scale \citep{sawada2022concept, yun2023do, yuksekgonul2023post}. 
%%% recent work with LLMs
To address this, recent work automates concept construction using either LLMs to generate class-specific semantic descriptions~\citep{yang2023language, oikarinen2023label, yan2023learning, srivastava2024vlg}, or VLMs to select visual concepts based on image-text alignment~\citep{rao2024discover, he2025v2c}.
%%% challenges
These methods reduce annotation effort and perform competitively at scale. 
However, as summarized in Table~\ref{tab:comparison_single}, they still fall short in addressing three fundamental challenges:
\begin{itemize}
  \item \textbf{Poor Visual Grounding.}
  LLM-generated concepts offer semantic richness but often lack alignment with actual visual content~\citep{yang2023language, oikarinen2023label}. 
  Conversely, VLM-based methods improve visual fidelity but sacrifice class-level semantic coherence and incur higher computational costs~\citep{rao2024discover, he2025v2c}. 
  It reflects a persistent semantic-visual gap that weakens both accuracy and interpretability.

  \item \textbf{Concept Redundancy.} 
  As depicted in Figure \ref{fig:teaser}(a) and Figure \ref{fig:teaser}(b), some methods generate concepts independently for each class, resulting in semantic duplication and overlapping terms~\citep{yang2023language, srivastava2024vlg, he2025v2c}; Others adopt global deduplication, which compresses redundancy but forces unrelated classes to share a fixed pool of concepts, undermining class discrimination~\citep{oikarinen2023label, yuksekgonul2023post, shang2024incremental}.
  Both strategies hinder model clarity and training stability.

  \item \textbf{Inadequate Metrics.} 
  Most CBMs are evaluated solely on classification accuracy, ignoring the interpretability cost of large and redundant concept sets~\citep{yang2023language, rao2024discover, he2025v2c, prasse2025dcbm}.
  As shown in Table~\ref{tab:comparison_single}, few models address this concern explicitly. Without principled metrics to capture the trade-off between accuracy and concept efficiency, performance gains may come at the expense of usability.
\end{itemize}

%%% our contributions: PS-CBM
To address these challenges, we propose \textbf{PS-CBM}, a unified framework for interpretable and scalable \textbf{C}oncept \textbf{B}ottleneck \textbf{M}odeling based on a novel \textbf{P}artially \textbf{S}hared concept strategy.
Our core contributions are as follows:
\begin{itemize}
  \item \textbf{Multimodal Concept Generation.} 
  We integrate the semantic richness of LLMs with visual grounding from exemplar images, generating concept sets that are both semantically meaningful and visually faithful, bridging the semantic-visual gap.
  
  \item \textbf{Partially Shared Concept Strategy.} 
  We introduce a novel strategy that merges concepts with similar activation patterns and assigns them across all relevant classes. 
  As depicted in Figure \ref{fig:teaser}(c), this partially shared strategy combines the specificity of per-class concepts with the compactness of global sharing, reducing redundancy without sacrificing discriminative expressiveness.
  
  \item \textbf{Concept-Efficient Accuracy (CEA).} 
  We propose a task-aware, post-hoc metric that jointly quantifies classification accuracy and concept compactness. 
  CEA is interpretable, bounded, model-agnostic, and requires no changes to model training, enabling fair comparison across CBM designs.
\end{itemize}

As shown in Table~\ref{tab:comparison_single}, unlike previous approaches that only partially (or fail to) address these challenges, PS-CBM achieves comprehensive coverage across all dimensions (indicated by $\checkmark$), showcasing its robustness and design coherence.
%%% experiments
To validate the effectiveness of PS-CBM, we conduct extensive experiments across eleven diverse real-world datasets, covering a broad spectrum of classification tasks, from general-purpose to fine-grained and domain-specific challenges.
%%%
PS-CBM consistently surpasses state-of-the-art CBMs in classification accuracy (+1.0\%--7.4\%) and
CEA (+2.0\%--9.5\%). 
More importantly, it does so with significantly fewer concepts, underscoring its ability to deliver high predictive performance while preserving transparency. 
% These results affirm PS-CBM as a practical and principled solution for interpretable image classification at scale.

\section{Related Work}
\label{sect:related_work}

\begin{figure*}[t]
  \centering
  \includegraphics[width=0.99\textwidth]{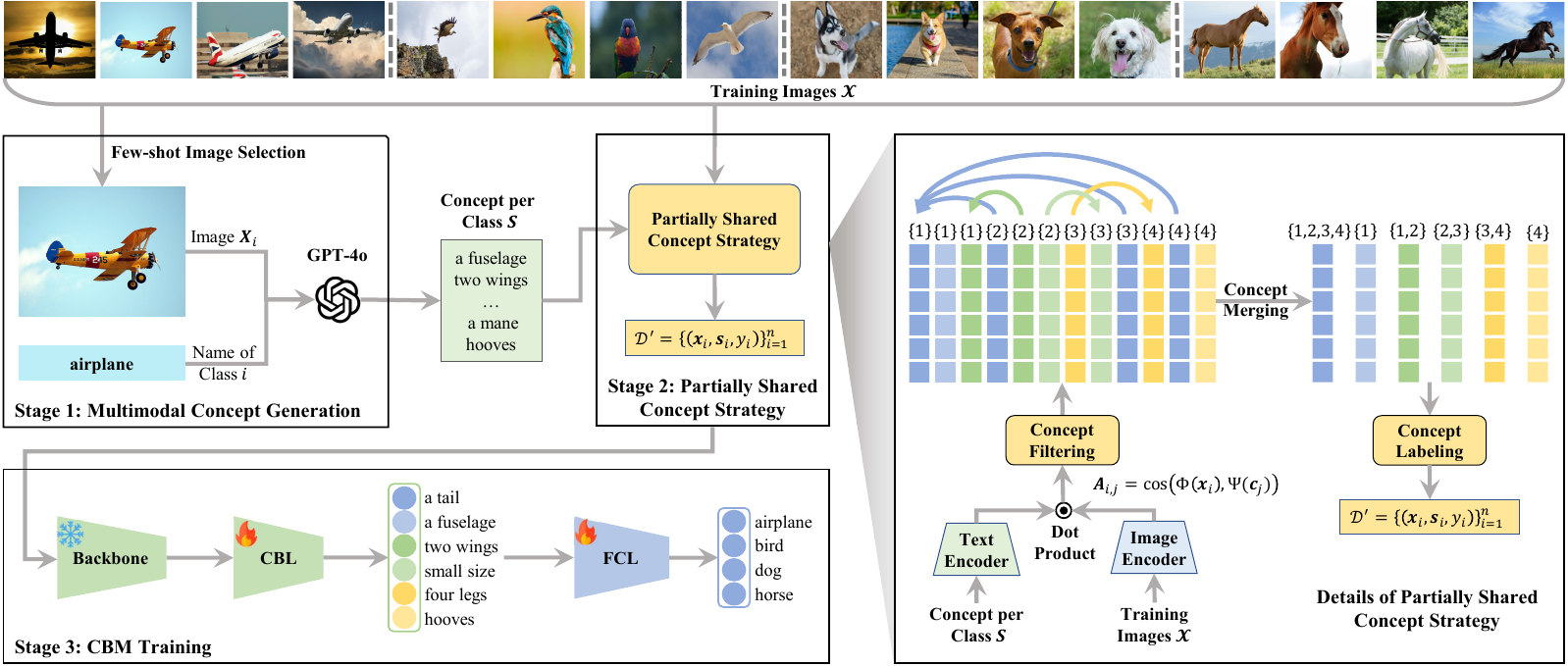}
  \caption{Overview of the PS-CBM pipeline. Stage 1: Generate multimodal concepts $\mathcal{S}$ by aligning LLM-derived semantics with exemplar images. Stage 2: Apply the Partially Shared Concept Strategy based on activation patterns to construct a concept-labeled dataset $\mathcal{D}'$. Stage 3: Train a transparent sequential predictor on $\mathcal{D}'$ for interpretable image classification.}
  \label{fig:framework}
\end{figure*}

\subsection{Concept Bottleneck Models}
\label{sect:related_work:cbm}

Concept Bottleneck Models (CBMs) enhance model interpretability by introducing human-understandable concepts as an intermediate representation between inputs and outputs~\citep{koh2020concept}. 
Existing approaches largely fall into two categories: those using a \textbf{globally shared} concept pool and those employing \textbf{independent}, class-specific pools.

\paragraph{Globally Shared Concept Pools}
These methods use a unified concept set shared across \emph{all} classes, enabling reusability and scalability \citep{oikarinen2023label, yuksekgonul2023post, shang2024incremental, rao2024discover, midavaine2024re, prasse2025dcbm, ruiz2024theoretical, panousis2024coarse, vandenhirtz2024stochastic, penaloza2025addressing, zarlenga2025avoiding, schmalwasser2025fastcav, xie2025discovering}.
%%%
For instance, \textbf{LF-CBM} \citep{oikarinen2023label} generates class-level concepts using GPT-3 \citep{brown2020language} to reduce annotation cost, while \textbf{Res-CBM} \citep{shang2024incremental} incrementally expands the concept space through residual learning.
\textbf{DN-CBM} \citep{rao2024discover} employs sparse autoencoders to discover transferable visual concepts, and \textbf{DCBM} \citep{prasse2025dcbm} extracts multi-granular concepts via segmentation foundation models.
%%%
Despite their scalability, these methods often suffer from \emph{concept redundancy}, where irrelevant or overlapping concepts impair discriminative capacity and clarity.

\paragraph{Independent Concept Pools}
Alternatively, class-specific models such as \textbf{LaBo} \citep{yang2023language}, \textbf{VLG-CBM}~\citep{srivastava2024vlg}, and \textbf{V2C-CBM} \citep{he2025v2c} generate tailored concepts for each class, enhancing semantic relevance.
However, this design introduces \emph{semantic duplication}, where similar concepts are redundantly assigned to multiple classes, leading to inefficiency and potential information leakage.
%%%
To overcome the limitations of both extremes, we introduce a \textbf{Partially Shared Concept Strategy} that adaptively merges semantically similar concepts, balancing compactness and discriminative power.

\paragraph{Alternative Interpretability Strategies}
Beyond concept pooling, other methods explore interpretability through different mechanisms~\citep{yan2023learning, delfosse2024interpretable, bhalla2024interpreting, laguna2024beyond, tan2024explain, liu2024concept, huang2024concept, dominici2025counterfactual, benou2025show, yu2025language, penaloza2025addressing, hu2025editable, liu2025hybrid, yamaguchi2025explanation}.
For example, \textbf{LM4CV}~\citep{yan2023learning} uses LLMs to retrieve image-specific concepts, but lacks consistent concept-class mappings, limiting coherence and interpretability.
%%%
\textbf{XBM}~\citep{yamaguchi2025explanation} generates textual explanations from embeddings using LLMs. While expressive, it lacks concept-level transparency and requires large backbones, making it impractical for lightweight models such as ResNet50~\citep{he2016deep}.
Recently, \textbf{Chat-CBM}~\citep{he2025chat} enhances human intervention through language-driven editing, yet concept redundancy and weak grounding remain open challenges.

\subsection{Metrics for Concept-based Models}
\label{sect:related_work:metrics}

CBMs are typically evaluated by accuracy alone, often overlooking the cost of large or redundant concept sets.
%%%
To address this, \textbf{Concept Utilization Efficiency (CUE)}~\citep{shang2024incremental} penalizes verbose concept sets but lacks a clear upper bound and is sensitive to textual formatting.
%%%
\textbf{Number of Effective Concepts (NEC)}~\citep{srivastava2024vlg} quantifies concept sparsity but requires training-time changes and hyperparameter tuning.
%%%
We propose \textbf{Concept-Efficient Accuracy (CEA)}, a post-hoc, task-aware metric that balances accuracy and concept compactness.
CEA is text-invariant, training-free, and enables fair comparison across different CBM architectures.

\section{The PS-CBM Framework}
\label{sect:framework}

We propose \textbf{PS-CBM}, a unified framework for constructing interpretable concept bottlenecks via a three-stage pipeline:
\begin{enumerate}
  \item \textbf{Multimodal Concept Generation}, which leverages both language and vision modalities to produce semantically meaningful and visually grounded concepts;
  \item \textbf{Partially Shared Concept Strategy}, which adaptively assigns concepts to classes by merging semantically overlapping concepts based on activation patterns; 
  \item \textbf{CBM Training}, which learns a transparent prediction model through concept supervision.
\end{enumerate}
The overall architecture is illustrated in Figure~\ref{fig:framework}.

\paragraph{Problem Setup}
Let $\mathcal{D} = \{(\bm{x}_i, y_i)\}_{i=1}^n$ be a dataset with $n$ samples, where $\bm{x}_i$ is an image from class $y_i \in \mathcal{Y} = \{1, \cdots, l\}$.
%%%
Each class $i$ has its own image set $\mathcal{X}_i$, with $\mathcal{X} = \bigcup_{i=1}^l \mathcal{X}_i$, and candidate concept set $\mathcal{S}_i$, with $\mathcal{S} = \bigcup_{i=1}^l \mathcal{S}_i$ and $|\mathcal{S}| = m$.
%%%
The goal is to learn a classification function:
\begin{displaymath}
  \bm{f} \circ \bm{g} \circ \bm{\phi}: \mathcal{X} \rightarrow \mathcal{Y},
\end{displaymath}
where predictions are mediated through a binary concept space for interpretability.

\subsection{Multimodal Concept Generation}
\label{sect:framework:Concept_generate}

We generate interpretable and grounded concepts using both semantic prompts and exemplar images.

\paragraph{Few-shot Image Selection} 
To anchor concepts visually, we construct a diverse, few-shot exemplar set $\bm{X}_i \subset \mathcal{X}_i$ for each class $i$ ($1 \leq i \leq l$) using CLIP embedding \citep{radford2021learning}.
%%%
Specifically, we initialize with a random image and iteratively select additional exemplars that maximize cosine distance from those already selected.
%%%
For noisy datasets (e.g., Food101 \citep{bossard2014food}), we employ random sampling.

\paragraph{Concept Generation}
For each class $i$, we construct a prompt by combining a text description with the selected exemplars $\bm{X}_i$ and query GPT-4o twice to reduce randomness. 
%%%
After deduplication, we obtain a candidate concept set $\mathcal{S} = \bigcup_{i=1}^l \mathcal{S}_i$. 
Each concept $\bm{c}_j$ ($1 \leq j \leq m$) is associated with a class set $\mathcal{C}_j$.

\subsection{Partially Shared Concept Strategy}
\label{sect:framework:shared_concept}

To reduce redundancy and enhance interpretability, we refine $\mathcal{S}$ in three steps:

\paragraph{Step 1: Concept Filtering}
Let $\Phi(\cdot)$ and $\Psi(\cdot)$ denote the image and text encoders, respectively.
We compute the affinity matrix $\bm{A} \in \mathbb{R}^{n \times m}$ between images and concepts:
\begin{displaymath}
  \bm{A}_{i,j} = \cos(\Phi(\bm{x}_i), \Psi(\bm{c}_j)).
\end{displaymath}
We retain concept $\bm{c}_j$ if its average top-4 alignment with class images exceeds the confidence threshold $\tau_{\text{conf}}$.

\begin{algorithm}[t]
\caption{Concept Merging}
\label{alg:concept_merging}
\KwIn{filtered concepts $\bm{S}$; filtered affinity matrix $\bm{A}$; merge threshold $\tau_{\text{merge}}$;}
\KwOut{The final concept set $\hat{\mathcal{S}}$ after merging;}
\BlankLine
$m = |\mathcal{S}|$\;
\For{$i=1$ \KwTo $m$}{
  \For{$j=1$ \KwTo $m$}{
    $\bm{Q}_{i,j} = \frac{\bm{A}_{:,i}^\top \bm{A}_{:,j}}{\|\bm{A}_{:,i}\|\|\bm{A}_{:,j}\|}$\;
  }
}
$\hat{\mathcal{S}} \gets \emptyset$\;
\While{$\mathcal{S} \neq \emptyset$}{
  $\mathcal{S}_j \gets \{ \bm{c}_i \in \mathcal{S} \mid \bm{Q}_{i,j} > \tau_{\text{merge}}\}$ \textbf{foreach} $\bm{c}_j \in \mathcal{S}$\; 
  Retrieve $\bm{c}_{\text{max}} \gets \argmax_{\bm{c}_j \in \mathcal{S}} |\mathcal{S}_j|$ and $\mathcal{S}_{\text{max}}$\;
  
  $\hat{\mathcal{S}} \gets \hat{\mathcal{S}} \cup \{\bm{c}_{\text{max}}\}$\;
  $\mathcal{S} \gets \mathcal{S} \setminus (\{\bm{c}_{\text{max}}\} \cup \mathcal{S}_{\text{max}})$\;
  % $\mathcal{C}_r \leftarrow \bigcup_{\bm{c}_j \in \mathcal{R}_r} \mathcal{C}_j$\;
}
\Return $\hat{\mathcal{S}}$\;
\end{algorithm}
\setlength{\textfloatsep}{1.0em}

\paragraph{Step 2: Concept Merging}
Algorithm~\ref{alg:concept_merging} outlines the concept merging process. 
We begin by computing a correlation matrix $\bm{Q}$ over filtered concepts (Lines 1--4) and greedily merge those exceeding a threshold $\tau_{\text{merge}}$) (Lines 5--11).
%%%
Merged concepts $\hat{\mathcal{S}}$ inherit the union of original class sets.
%%%
To limit redundancy, we retain only the top $K$ exclusive concepts per class, i.e., those associated with a single class.

\paragraph{Step 3: Concept Labeling}
Let $\hat{m} = |\hat{\mathcal{S}}|$. 
The one-hot encoded concept label vector $\bm{s}_i = [s_{i,j}] \in \{0,1\}^{\hat{m}}$ for each image $\bm{x}_i$ is defined as:
\begin{displaymath}
  s_{i,j} = \begin{cases}
    1, & \text{if } y_i \in C_j~\text{and}~\bm{A}_{i,j} > \tau_{\text{conf}}, \\
    0, & \text{otherwise}.
  \end{cases}
\end{displaymath}
This yields the concept-labeled dataset  $\mathcal{D}'$ for training CBM:
\begin{displaymath}
  \mathcal{D}' = \{ (\bm{x}_i, \bm{s}_i, y_i) \}_{i=1}^n.
\end{displaymath}

\begin{table*}[t]
\centering
\small
\setlength\tabcolsep{3pt}
\renewcommand{\arraystretch}{1.3}
\resizebox{0.99\textwidth}{!}{%
\begin{tabular}{l *{11}{cc}}
\toprule
\multirow{2.5}{*}{\textbf{Method}} 
& \multicolumn{2}{c}{\textbf{Aircraft}}
& \multicolumn{2}{c}{\textbf{CIFAR10}}
& \multicolumn{2}{c}{\textbf{CIFAR100}}
& \multicolumn{2}{c}{\textbf{CUB200}}
& \multicolumn{2}{c}{\textbf{DTD}}
& \multicolumn{2}{c}{\textbf{Flower102}}
& \multicolumn{2}{c}{\textbf{Food101}}
& \multicolumn{2}{c}{\textbf{HAM10000}}
& \multicolumn{2}{c}{\textbf{ImageNet}}
& \multicolumn{2}{c}{\textbf{Resisc45}}
& \multicolumn{2}{c}{\textbf{UCF101}} \\
\cmidrule(lr){2-3} \cmidrule(lr){4-5} \cmidrule(lr){6-7} \cmidrule(lr){8-9} \cmidrule(lr){10-11} 
\cmidrule(lr){12-13} \cmidrule(lr){14-15} \cmidrule(lr){16-17} \cmidrule(lr){18-19} \cmidrule(lr){20-21} \cmidrule(lr){22-23}
& \textbf{ACC $\uparrow$} & \textbf{CEA $\uparrow$} 
& \textbf{ACC $\uparrow$} & \textbf{CEA $\uparrow$} 
& \textbf{ACC $\uparrow$} & \textbf{CEA $\uparrow$} 
& \textbf{ACC $\uparrow$} & \textbf{CEA $\uparrow$} 
& \textbf{ACC $\uparrow$} & \textbf{CEA $\uparrow$} 
& \textbf{ACC $\uparrow$} & \textbf{CEA $\uparrow$} 
& \textbf{ACC $\uparrow$} & \textbf{CEA $\uparrow$} 
& \textbf{ACC $\uparrow$} & \textbf{CEA $\uparrow$} 
& \textbf{ACC $\uparrow$} & \textbf{CEA $\uparrow$} 
& \textbf{ACC $\uparrow$} & \textbf{CEA $\uparrow$} 
& \textbf{ACC $\uparrow$} & \textbf{CEA $\uparrow$} \\
\midrule
\textbf{Linear Probe} 
& 42.5 & -- & 88.5 & -- & 69.8 & -- & 67.4 & -- & 71.8 & -- & 97.4 & -- 
& 82.7 & -- & 79.8 & -- & 72.6 & -- & 85.4 & -- & 81.0 & -- \\
\midrule
\textbf{LaBo} 
& 40.3 & 27.9 & 87.5 & 60.1 & 68.1 & 47.1 & 66.8 & 46.1 & 71.3 & 49.4 & 96.6 & 66.7 
& 82.2 & 56.8 & 77.1 & 50.7 & 72.1 & 49.0 & 84.1 & 58.4 & 79.9 & 55.2 \\
\textbf{LF-CBM}
& 36.2 & 28.9 & 87.3 & 63.1 & \underline{68.8} & 49.9 & 58.6 & 45.7 & 68.5 & 52.8 & 94.4 & \underline{73.1} 
& 77.7 & 58.8 & 70.2 & 56.8 & 67.5 & 48.8 & 84.5 & 62.2 & 80.0 & 58.2 \\
\textbf{LM4CV} 
& 38.5 & 28.8 & 81.4 & 62.8 & 65.9 & 49.3 & 65.6 & 48.6 & 70.8 & 53.6 & 94.6 & 70.7 
& 81.1 & 60.6 & 66.8 & 49.8 & 69.6 & 50.2 & 81.3 & 61.7 & 78.9 & 59.0 \\
\textbf{DN-CBM}
& 42.1 & 28.7 & 88.2 & 55.2 & 68.6 & 46.7 & 66.6 & 46.2 & \underline{74.5} & 49.8 & 96.6 & 65.8 
& \underline{82.3} & 56.1 & \underline{80.5} & 47.6 & \underline{73.1} & 52.0 & \underline{85.7} & 57.3 & 81.1 & 55.3 \\
\textbf{Res-CBM} 
& 36.9 & 28.9 & 87.6 & 61.9 & 65.7 & 49.6 & 59.1 & 46.9 & 64.4 & 49.5 & 93.8 & 73.1 
& 79.3 & \underline{61.6} & 77.5 & 52.7 & 68.3 & \underline{52.2} & 81.8 & 61.5 & 75.4 & 58.6 \\
\textbf{VLG-CBM} 
& \underline{45.6} & \underline{34.6} & \underline{89.6} & \underline{67.4} & 68.3 & \underline{50.8} & \underline{68.0} & \underline{51.0} & 72.2 & \underline{55.3} & \underline{97.1} & 72.5 
& 81.6 & 60.3 & 79.8 & \underline{59.9} & 65.7 & 47.6 & 84.9 & \underline{65.1} & \underline{81.3} & \underline{60.8} \\
\textbf{V2C-CBM}
& 37.1 & 25.6 & 87.3 & 60.0 & 68.3 & 47.2 & 63.8 & 44.0 & 70.8 & 49.1 & 96.6 & 68.2 
& 81.2 & 56.1 & 75.4 & 49.6 & 73.0 & 49.6 & 83.7 & 58.1 & 78.3 & 54.1 \\
\textbf{DCBM} 
& 36.4 & 25.8 & 85.7 & 55.9 & 62.3 & 44.3 & 59.8 & 43.2 & 70.1 & 48.8 & 94.0 & 66.8 
& 79.3 & 56.4 & 75.3 & 46.4 & 57.0 & 42.3 & 81.1 & 56.4 & 75.3 & 53.6 \\
\textbf{PS-CBM} 
& \textbf{47.0} & \textbf{34.9} & \textbf{89.8} & \textbf{68.5} & \textbf{72.1} & \textbf{53.6} & \textbf{70.1} & \textbf{53.3} & \textbf{75.1} & \textbf{59.0} & \textbf{97.9} & \textbf{74.9} 
& \textbf{83.0} & \textbf{61.7} & \textbf{83.4} & \textbf{61.3} & \textbf{74.0} & \textbf{54.6} & \textbf{87.5} & \textbf{65.7} & \textbf{83.0} & \textbf{61.8} \\
\bottomrule
\end{tabular}}
\caption{ACC and CEA on 11 datasets using \texttt{CLIP\_RN50} backbone. \textbf{Bold} and \underline{underline} denote the best and second-best results.}
\label{tab:acc_cea_results}
\end{table*}

\begin{figure*}[!t]
  \centering
  \includegraphics[width=0.99\textwidth]{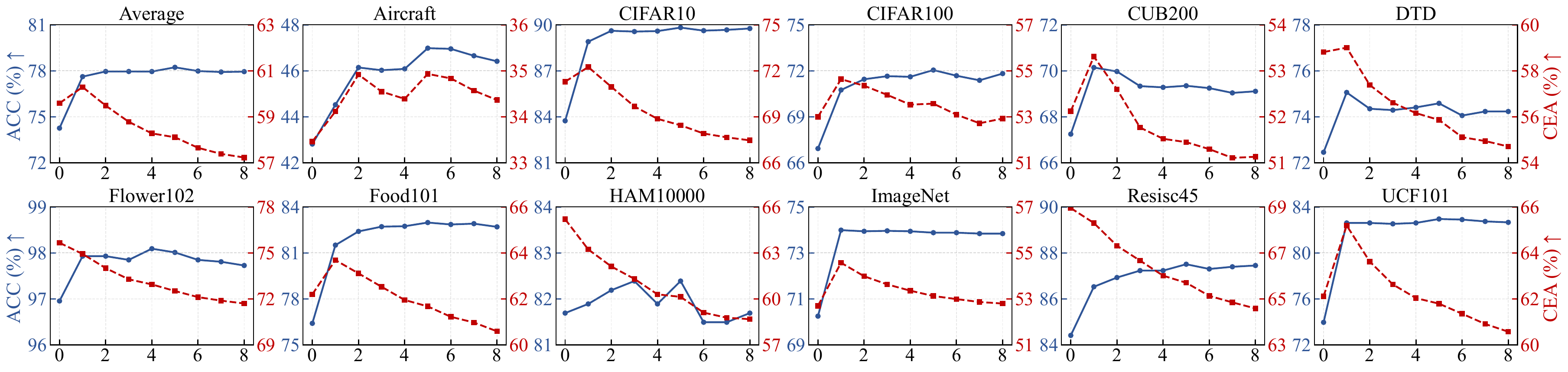}
  \caption{Variations of ACC and CEA with the upper limit $K$ on the number of exclusive concepts per class across datasets.}
  \label{fig:Acc_CEA_vs_Kindep}
\end{figure*}

\subsection{CBM Training}
\label{sect:framework:train_cbm}

\paragraph{Training Concept Bottleneck Layer (CBL)}
Given the concept-labeled dataset $\mathcal{D}'$, we train the CBL to predict multi-label concept annotations. 
Let $\bm{\phi}: \mathcal{X} \to \mathbb{R}^d$ denote a \emph{frozen} backbone encoder mapping each image $\bm{x}$ to an embedding $\bm{z} = \bm{\phi}(\bm{x})$. 
%%%
The CBL $\bm{g}: \mathbb{R}^d \to \mathbb{R}^{\hat{m}}$ projects embeddings to concept logits. 
We optimize $\bm{g}$ by minimizing Binary Cross-Entropy (BCE) loss:
\begin{equation}
\label{eq:CBL}
  \textstyle \min_{\bm{g}} \mathcal{L}_{\text{CBL}} = \frac{1}{n} \sum_{i=1}^{n} \text{BCE}(\bm{g}(\bm{\phi}(\bm{x}_i)), \bm{s}_i), % \textstyle 
\end{equation}
where $\bm{s}_i$ denotes a vector of the binary concept labels.

\paragraph{Training Final Classification Layer (FCL)}
At last, we train a sparse linear classifier $\bm{f}: \mathbb{R}^{\hat{m}} \to \mathbb{R}^l$ with weight matrix $\bm{W}_F$ and bias $\bm{b}_F$, to map concept logits to class predictions.
For each sample, we first compute concept logits via the trained, frozen CBL $\bm{g}$ and normalize them using training set statistics. 
The optimization objective combines Cross-Entropy (CE) loss and elastic-net regularization~\citep{zou2005regularization}:
\begin{equation}
\label{eq:final-layer}
  \textstyle  \min_{\bm{f}} \mathcal{L}_{\text{FCL}} = \frac{1}{n}\sum_{i=1}^{n} \text{CE}(\bm{f}(\hat{\bm{g}}(\bm{x}_i)),y_i) + \lambda R_{\alpha}(\bm{W}_F), % \textstyle 
\end{equation}
where $\hat{\bm{g}}(\bm{x}_i)$ denotes the normalized concept logits, and
\begin{displaymath}
  R_{\alpha}(\bm{W}_F) = (1-\alpha)\|\bm{W}_F\|_2^2 + \alpha \|\bm{W}_F\|_1.
\end{displaymath}
We solve this objective using the GLM-SAGA \citep{zou2005regularization} optimizer.

%\subsection{Concept-Efficient Accuracy (CEA)}
%\label{sect:framework:cea}

%Finally, we propose CEA as a metric to quantify the trade-off between predictive accuracy and interpretability.
%%%
%Let $m$ be the number of concepts used and $k = \lceil \log_2 l \rceil$ represent the information-theoretic minimum required to distinguish among $l$ classes. We define CEA as:

%\begin{equation}
%  \text{CEA} = \tfrac{\mathrm{ACC}}{(\log_k m)^{\beta}},
%\end{equation}
%where $\mathrm{ACC} \in [0,1]$ denotes the model's classification accuracy, and $\beta \ge 0$ is a temperature controlling the penalty on concept complexity. 
%%%
%A smaller $\beta$ makes CEA focus more on accuracy, while a larger $\beta$ prioritizes concept compactness.
%%%
%CEA possesses three desirable properties:
%\begin{itemize}
%  \item \textbf{Optimal Efficiency}: CEA approaches 1 as accuracy nears 1 ($\text{ACC} \to 1$) and concept usage approaches the theoretical minimum ($m \to k$).

%  \item \textbf{Adaptive Scaling}: Base‑$k$ logarithmic scaling ensures the penalty adapts to task complexity: more classes allow moderately larger concept sets without excessive penalty.

%  \item \textbf{Theoretical Foundation}: The formulation aligns with the Minimum Description Length principle \cite{rissanen1978modeling, rissanen1986stochastic, tishby2000information}, encouraging parsimonious yet accurate explanations.
%\end{itemize}
%In summary, CEA provides a unified measure of both accuracy and interpretability, supporting fair, task-aware comparisons across CBMs with varying concept complexities.

\subsection{Concept-Efficient Accuracy (CEA)}
\label{sect:framework:cea}
CEA is grounded in Shannon’s information theory, which establishes that distinguishing among $l$ classes using binary (0/1) concept signals requires at least $k = \lceil \log_2 l \rceil$ bits of information~\cite{shannon1948mathematical}.
CEA then quantifies how efficiently a model achieves its accuracy relative to this theoretical bound, penalizing redundant concept usage. 
%%%
Let $m$ denote the number of concepts used. We define CEA as:
\begin{equation}
  \textstyle \text{CEA} = \frac{\mathrm{ACC}}{(\log_k m)^{\beta}},
\end{equation}
where $\mathrm{ACC} \in [0,1]$ denotes the model’s classification accuracy, and $\beta \ge 0$ is a temperature parameter controlling the penalty on concept complexity. 
A smaller $\beta$ makes CEA focus more on accuracy, whereas a larger $\beta$ emphasizes concept compactness.
CEA possesses three desirable properties:
\begin{itemize}
  \item \textbf{Optimal Efficiency}: CEA approaches 1 as accuracy nears 1 ($\mathrm{ACC} \to 1$) and concept usage approaches the theoretical minimum ($m \to k$).

  \item \textbf{Adaptive Scaling}: The base-$k$ logarithmic scaling ensures that the penalty adapts to task complexity: more classes allow moderately larger concept sets without excessive penalization.

  \item \textbf{Theoretical Foundation}: The formulation aligns with Shannon’s information theory, encouraging parsimonious yet accurate explanations.
\end{itemize}

In summary, CEA provides a unified measure of both accuracy and interpretability, enabling fair, task-aware comparisons across CBMs with varying concept complexities.

\section{Experiments}
\label{sect:expt}

We conduct extensive experiments to evaluate PS-CBM's classification performance, interpretability, and robustness. % across diverse visual domains.
%%%
% Our study addresses the following research questions:
% \begin{itemize}
%   \item Does PS-CBM maintain high accuracy while reducing concept complexity? (Section \ref{sect:expt:classification})
%   \item How does PS-CBM compare with state-of-the-art CBMs under interpretability metrics? (Section \ref{sect:expt:explanability})
%   \item What is the contribution of each component in PS-CBM through ablations? (Section \ref{sect:expt:ablation})
% \end{itemize}

\begin{figure*}[t]
  \centering
  \includegraphics[width=0.99\textwidth]{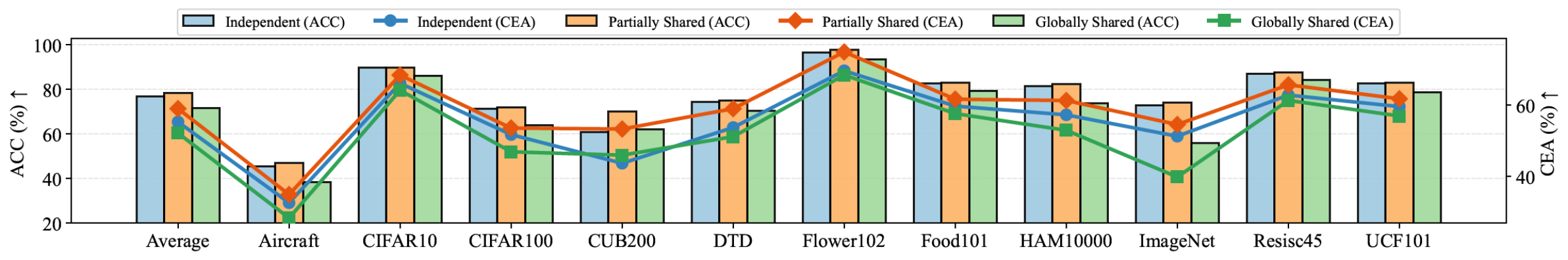}
  \caption{Ablation study on different concept bottleneck strategies, comparing their impact on ACC and CEA.}
  \label{fig:ablation_different_strategies}
\end{figure*}

\begin{figure*}[t]
  \centering
  \includegraphics[width=0.99\textwidth]{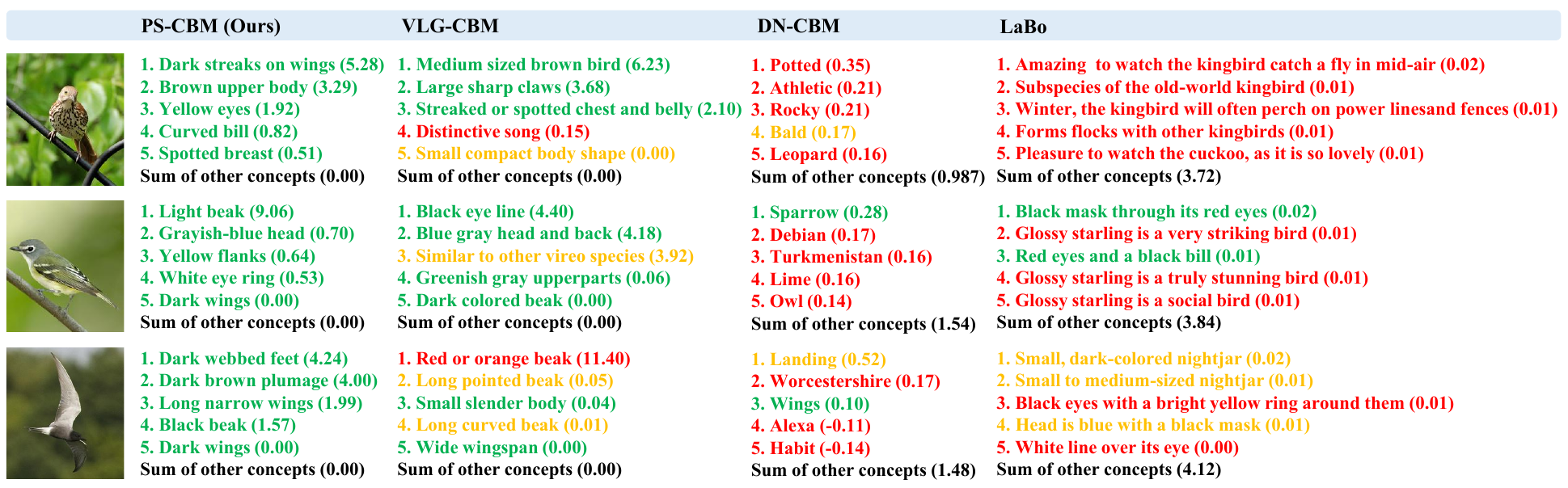}
  \caption{Case study of PS-CBM predictions. Green highlights correct predictions, yellow denotes partially accurate or ambiguous outputs, and red indicates incorrect results.}
  \label{fig:case_study}
\end{figure*}

\subsection{Experimental Setup}
\label{sect:expt:setup}

\paragraph{Datasets}
We evaluate PS-CBM on 11 publicly available real-world datasets spanning multiple domains:
(1) General image classification: \textbf{CIFAR10}, \textbf{CIFAR100}~\cite{krizhevsky2009learning}, and  \textbf{ImageNet}~\cite{deng2009imagenet}; 
(2) Fine-grained classification: \textbf{Aircraft}~\cite{maji2013fine}, \textbf{CUB200}~\cite{wah2011caltech}, \textbf{Flower102}~\cite{nilsback2008automated}, and \textbf{Food101}~\cite{bossard2014food}; 
(3) Domain-specific tasks: \textbf{DTD}~\cite{cimpoi2014describing} (textures), \textbf{HAM10000}~\cite{tschandl2018ham10000} (skin tumor classification), \textbf{Resisc45}~\cite{cheng2017remote} (remote sensing), and \textbf{UCF101}~\cite{soomro2012ucf101} (action recognition).

\paragraph{Metrics}
We evaluate performance using three key metrics:
\begin{itemize}
  \item \textbf{Classification Accuracy (ACC):} Serving as the standard measure of predictive performance.

  \item \textbf{Concept-Efficient Accuracy (CEA):} Our proposed metric balancing accuracy and concept compactness.

  \item \textbf{CLIP Score (in ablations):} Assessing concept–image alignment, especially for domain-specific datasets like DTD, Resisc45, and UCF101, where it is crucial.
\end{itemize}

\paragraph{Baselines}
We assess PS-CBM against two categories of models.
%%%
The first comprises leading CBMs, including \textbf{LaBo}~\cite{yang2023language}, \textbf{LF-CBM}~\cite{oikarinen2023label}, \textbf{LM4CV}~\cite{yan2023learning}, \textbf{DN-CBM}~\cite{rao2024discover}, \textbf{Res-CBM}~\cite{shang2024incremental}, \textbf{VLG-CBM}~\cite{srivastava2024vlg}, \textbf{V2C-CBM}~\cite{he2025v2c}, and \textbf{DCBM}~\cite{prasse2025dcbm}.
%%%
The second is the \textbf{Linear Probe}~\cite{yang2023language}, a logistic regression model trained on CLIP features, serving as a strong black-box baseline without explicit interpretability.

\paragraph{Implementation Details}
All methods share identical train/dev/test splits and backbones (\texttt{CLIP\_RN50} and \texttt{CLIP\_ViT-L/14}); results for the latter appear in Appendix B. 
Image-text similarities use \texttt{CLIP\_ViT-B/16}.
%%%
Concept filtering and labeling uses $\tau_{\text{conf}} = 0.20$. 
Concept merging uses $\tau_{\text{merge}} \in [0.9996, 0.9999]$ (step 0.0001), and the pruning parameter $K$ from $\{0,1,\cdots,8\}$. 
%%%
The CEA temperature parameter $\beta$ is set to $0.25$.
Due to ImageNet's scale, only 10\% of its training images are sampled for merging. 
Full configurations are provided in Appendix A.

\paragraph{Experiment Environment}
All experiments are run on Ubuntu 22.04 with PyTorch 2.3.0, CUDA 12.1, and a single NVIDIA RTX 3080 Ti (12GB), using an Intel\textsuperscript{\textregistered} Xeon\textsuperscript{\textregistered} 4214R CPU (12-core, 2.40 GHz) and 90 GB RAM.

\subsection{Classification Accuracy}
\label{sect:expt:classification}

Table \ref{tab:acc_cea_results} presents the classification accuracy (ACC) of PS-CBM and all baseline models across 11 datasets. 
%%%
PS-CBM consistently achieves the highest accuracy, demonstrating robust generalization across a wide range of domains, including fine-grained, texture, and remote sensing tasks.
%%%
This result underscores PS-CBM's ability to maintain strong predictive performance even in challenging settings.

To further contextualize PS-CBM's performance, Table \ref{tab:average_metric_baseline} presents the average classification accuracy alongside the number of concepts used by each method. 
PS-CBM surpasses state-of-the-art CBMs by 1.0\%--7.4\% in average ACC, while significantly reducing concept usage.
%%%
%Notably, it employs an average of 7,647 fewer concepts per dataset than DN-CBM--the closest competitor--underscoring its superior efficiency. 
%This substantial reduction not only enhances model interpretability but also mitigates risks associated with concept leakage.

Notably, PS-CBM uses an average of 7,647 fewer concepts per dataset than DN-CBM, achieving higher ACC and CEA. 
As LM4CV (Fig.~3) shows, large concept pools can inflate accuracy even with random concepts, indicating spurious correlations. PS-CBM's strong performance with fewer concepts demonstrates tighter concept grounding and reduced leakage risk.

\subsection{Explanability}
\label{sect:expt:explanability}

Alongside accuracy, Table \ref{tab:acc_cea_results} also reports CEA, a metric designed to capture the trade-off between predictive performance and interpretability. 
%%%
PS-CBM achieves the highest CEAs across all datasets, indicating that it delivers accurate predictions with a minimal and meaningful set of concepts.

\begin{table}[t]
\centering
\renewcommand{\arraystretch}{1.1}
\resizebox{0.99\linewidth}{!}{%
\begin{tabular}{lrrr}
\toprule
\textbf{Method} & \textbf{Avg. ACC (\%) $\uparrow$} & \textbf{Avg. \# Concepts $\downarrow$} & \textbf{Avg. CEA (\%) $\uparrow$} \\
\midrule
\textbf{LaBo}    & 72.8             & 7,900          & 51.6 \\
\textbf{LF-CBM}  & 72.9             & 718           & 55.2 \\
\textbf{LM4CV}   & 73.4             & 873           & 56.4 \\
\textbf{DN-CBM}  & \underline{77.3} & 8,192          & 53.4 \\
\textbf{Res-CBM} & 71.8             & \textbf{291}  & 56.7 \\
\textbf{VLG-CBM} & 75.2             & 732           & \underline{57.0} \\
\textbf{V2C-CBM} & 72.8             & 7,500          & 51.2 \\
\textbf{DCBM}    & 70.9             & 2,048          & 49.5 \\
\textbf{PS-CBM}  & \textbf{78.3} & \underline{545} & \textbf{59.0} \\
\bottomrule
\end{tabular}
}
\caption{Comparison of methods by average ACC, number of concepts used, and CEA. \textbf{Bold} and \underline{underlined} indicate the best and second-best results, respectively.}
\label{tab:average_metric_baseline}
\end{table}

Table \ref{tab:average_metric_baseline} presents the average CEA scores of all CBM methods across 11 datasets.
%%%
PS-CBM surpasses state-of-the-art CBMs by 2.0\%--9.5\% in average CEA, highlighting its ability to balance classification performance and concept compactness. 
%%%
In contrast, methods such as DN-CBM, LaBo, and V2C-CBM, despite achieving high accuracy, rely on excessively large concept sets, often an order of magnitude larger than others. 
This redundancy compromises interpretability and limits human oversight.
%%%
Their lower CEA values further emphasize the importance of jointly evaluating performance and interpretability, as captured by CEA.

\begin{table}[t]
\centering
\setlength\tabcolsep{3pt}
\renewcommand{\arraystretch}{1.15}
\resizebox{0.99\columnwidth}{!}{%
\begin{tabular}{llllll}
\toprule
\textbf{Method} & \textbf{DTD} & \textbf{Resisc45} & \textbf{UCF101} & \textbf{Concept Generation} & \textbf{Concept Pools} \\
\midrule
\textbf{LaBo}    & 0.227     & \underline{0.222} & 0.230 & Language & Independent\\
\textbf{LF-CBM}  & 0.225     & 0.208 & 0.199 & Language & Globally Shared\\
\textbf{DN-CBM}  & 0.192     & 0.187 & 0.187 & Vision & Globally Shared\\
\textbf{V2C-CBM} & \underline{0.246} & 0.216 & \underline{0.247} & Vision & Independent\\
\textbf{PS-CBM}  & \textbf{0.249} & \textbf{0.255} & \textbf{0.265} & Language+Vision & Partially Shared\\
\bottomrule
\end{tabular}%
}
\caption{CLIP Score ($\uparrow$) comparison across three domain-specific datasets. \textbf{Bold} and \underline{underlined} indicate the best and second-best results, respectively.}
% \caption{Comparison of different methods on special-domain datasets in terms of CLIP Score. \textbf{Bold} and \underline{underlined} denote the best and second-best results.}
\label{tab:clip_score_comparison}
\end{table}

To evaluate concept–image alignment, Table \ref{tab:clip_score_comparison} reports the average CLIP score at the class level. 
%%%
Compared to LaBo and LF-CBM, which use only LLMs for concept generation, and DN-CBM and V2C-CBM, which rely solely on visual alignment, PS-CBM achieves significantly better alignment on domain-specific datasets such as DTD, Resisc45, and UCF101. 
%%%
These results showcase the benefits of PS-CBM's multimodal concept generation strategy in producing concepts that are both semantically rich and visually grounded. %, particularly valuable in domains where textual and visual semantics diverge.

\subsection{Ablation Study}
\label{sect:expt:ablation}

To better understand the design choices in PS-CBM, we conduct several ablation studies.
%%%
Figure \ref{fig:Acc_CEA_vs_Kindep} illustrates how ACC and CEA vary with the maximum number of exclusive concepts per class ($K$). 
%%%
Accuracy improves sharply when $K$ increases from 0 to 1 and stabilizes beyond $K=2$.
Meanwhile, CEA peaks at $K=1$ and declines as $K$ increases. 
%%%
These results suggest that allowing a small number of class-specific concepts is beneficial, but excessive exclusivity undermines concept efficiency. 
This confirms that PS-CBM effectively leverages shared concepts while maintaining the discriminative power of a minimal set of exclusive ones.

Figure \ref{fig:ablation_different_strategies} compares three concept-sharing strategies: Independent, Partially Shared, and Globally Shared, implemented within the PS-CBM framework. 
%%%
The Partially Shared variant achieves the best performance in both ACC and CEA.
This validates the core idea of PS-CBM: selectively sharing concepts among semantically similar classes leads to more accurate and interpretable models than either fully disjoint or globally shared approaches.

\begin{table}[t]
\centering
\renewcommand{\arraystretch}{1.1}
\resizebox{0.99\columnwidth}{!}{
\begin{tabular}{lrrr}
  \toprule
  $\tau_{\text{conf}}$ & \textbf{Avg. ACC (\%) $\uparrow$} & \textbf{Avg. \# Concepts $\downarrow$} & \textbf{Avg. CEA (\%) $\uparrow$} \\
  \midrule
  0.10  & 76.20 & 548 & 57.41 \\
  0.15  & 76.14 & 548 & 57.36 \\
  0.20  & \textbf{78.35} & 545 & \textbf{59.02} \\
  0.25  & 72.71 & 458 & 55.16 \\
  0.30  & 57.55 & \textbf{145} & 46.84 \\
\bottomrule
\end{tabular}}
\caption{Effect of varying confidence threshold $\tau_{\text{conf}}$ on average ACC, number of concepts used, and CEA across all datasets. \textbf{Bold} highlights the best-performing setting.}
% \caption{Average accuracy, number of concepts used, and CEA at different values of confidence threshold $\tau_{\text{conf}}$. \textbf{Bold} denotes the best results.}
\label{tab:average_metric_confidence_threshold}
\end{table}

Table \ref{tab:average_metric_confidence_threshold} explores the impact of the confidence threshold $\tau_{\text{conf}}$ used in concept filtering and labeling, with thresholds ranging from 0.10 to 0.30. 
%%%
We observe that a threshold of 0.20 yields the best balance between ACC, \# Concepts, and CEA.
This result reinforces the robustness of PS-CBM under different hyperparameter settings.

\subsection{Case Study}
\label{sect:expt:case}

To further illustrate how PS-CBM supports interpretable decision-making, we present a case study in Figure \ref{fig:case_study} following the visualization protocol from VLG-CBM \citep{srivastava2024vlg}.
For each prediction, we compute the contribution of each concept based on its activation and corresponding weight, highlighting the top-5 concepts and aggregating the rest as ``sum of other concepts.''

Models with high interpretability tend to have a large share of their predictions explained by the top few concepts, making decisions easier to interpret and verify.
%%%
As shown in Figure \ref{fig:case_study}, PS-CBM and VLG-CBM exhibit more concentrated contributions from their top concepts compared to DN-CBM and LaBo.
%%%
In particular, PS-CBM benefits from both its multimodal concept generation and partially shared concept strategy, producing more relevant and semantically meaningful explanations. 
This leads to reduced concept redundancy and a lower risk of information leakage, thereby enhancing the overall transparency of the model.

\begin{figure}[t]
  \centering
  \includegraphics[width=0.99\columnwidth]{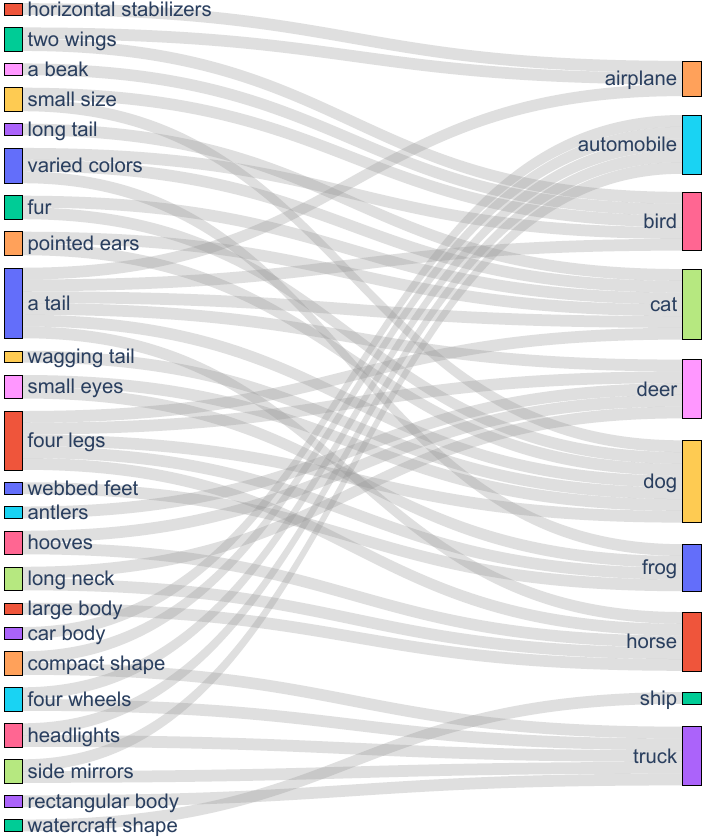}
  \caption{Concept-class map on CIFAR10 with $K=1$ and $\tau_{\text{merge}} = 0.9996$. Partially shared concepts appear across related classes, while others remain class-specific.}
  \label{fig:concept_class_map}
\end{figure}

\subsection{Effect of Partially Shared Concept Strategy}
\label{sect:expt:pscs}

To visualize the effect of the Partially Shared Concept Strategy (PSCS), we plot the concept-class map on CIFAR10 in Figure \ref{fig:concept_class_map}, using $K=1$ and a concept merge threshold of 0.9996. 
The visualization reveals that PSCS selectively shares concepts across semantically related classes while preserving class-specific ones where necessary.

For example, \textit{horizontal stabilizers} and \textit{a beak} appear only for airplanes and birds, respectively. 
%%%
Meanwhile, shared concepts such as \textit{two wings}, \textit{a tail}, and \textit{four legs} span related classes, reflecting common visual features.
%%%
Other shared attributes, such as \textit{hooves} or \textit{long neck}, apply to both deer and horses, while automotive-related features like \textit{headlights} and \textit{side mirrors} are shared between cars and trucks.

These patterns highlight PSCS's ability to learn compact, semantically coherent concepts, enhancing interpretability and robustness as a core contributor to PS-CBM's success.

\section{Conclusions}
\label{sect:conclusions}

%%% overall
In this paper, we introduced PS-CBM, a unified and scalable CBM framework for interpretable image classification. 
%%% technical novelty
PS-CBM introduces three key innovations: 
(1) a multimodal concept generation module that improves both relevance and interpretability; 
(2) a Partially Shared Concept Strategy that adaptively merges and reassigns concepts across classes based on activation patterns, effectively balancing specificity and compactness; and 
(3) a new metric, CEA, that jointly captures accuracy and concept efficiency.
%%% experiments
Extensive experiments on 11 diverse datasets show that PS-CBM consistently surpasses state-of-the-art CBMs in both accuracy and interpretability, while requiring significantly fewer concepts.
%%% impacts
Overall, PS-CBM offers a practical and extensible solution for building interpretable vision systems, laying a strong foundation for future research in scalable, multimodal, and concept-efficient learning.

\section*{Acknowledgements}
We sincerely thank the anonymous reviewers for their insightful and constructive feedback, which greatly improved this paper.
This work was supported by the National Natural Science Foundation of China (NSFC) under Grant Nos. 62125201 and U24B20174.

\bibliography{aaai2026.bib}
\appendix

\section{Implementation Details}
\label{appendix:impl}

\subsection{Dataset Details}
\label{appendix:impl:datasets}

To rigorously evaluate the performance, interpretability, and generalizability of PS-CBM, we conduct experiments across 11 publicly available datasets, which are selected to reflect a wide range of visual recognition challenges, including general-purpose object classification, fine-grained recognition, and domain-specific tasks such as medical diagnosis and satellite imagery analysis.
%%%
Below, we provide detailed descriptions and the unique characteristics of each dataset.

\paragraph{General Image Classification}
This category includes benchmarks designed to evaluate a model's ability to recognize common visual concepts across broad object categories. These datasets are commonly used in computer vision to validate generalization on everyday object classes:
\begin{itemize}
  \item \textbf{CIFAR10~\cite{krizhevsky2009learning}:} 
  A classic benchmark of 60,000 low-resolution (32$\times$32) color images evenly distributed across 10 object categories--airplane, automobile, bird, cat, deer, dog, frog, horse, ship, and truck. 
  The dataset provides 6,000 images per class, split into 50,000 for training and 10,000 for testing, making it a foundational testbed for general-purpose image classification.

  \item \textbf{CIFAR100~\cite{krizhevsky2009learning}:} 
  A more fine-grained variant of CIFAR10, this dataset contains the same number of total images (60,000) but subdivides them into 100 distinct classes grouped under 20 superclasses. Each class contains 600 images. 
  The increased granularity introduces additional complexity, making it a stronger benchmark for evaluating a model's discriminative capabilities across visually similar categories.

  \item \textbf{ImageNet~\cite{deng2009imagenet}:} 
  A large-scale benchmark composed of over 1.2 million high-resolution training images and 50,000 validation images, spanning 1,000 object categories. As one of the most influential datasets in deep learning, ImageNet has been instrumental in driving progress in image classification, representation learning, and transfer learning. Its scale, diversity, and high intra-class variability make it a robust test of generalization.
\end{itemize}

\paragraph{Fine-Grained Classification}
This task requires distinguishing between visually similar subcategories within a broader class (e.g., bird species, airplane models). 
Such datasets pose greater challenges due to subtle inter-class differences, requiring models to capture fine-grained and nuanced visual features.
\begin{itemize}
  \item \textbf{Aircraft~\cite{maji2013fine}:} 
  Comprising approximately 10,000 images categorized into 100 aircraft variants, this dataset requires models to differentiate fine-grained structural differences in aircraft such as engines, wings, and fuselage shape. 
  It poses a significant challenge due to inter-class similarity and intra-class variability caused by different angles, lighting, and environments.

  \item \textbf{CUB200~\cite{wah2011caltech}:} 
  Contains 11,788 images spanning 200 bird species, each annotated with detailed part locations (e.g., beak, wings, tail).
  It is a widely used benchmark in the fine-grained classification community, offering part-based and explainable modeling.
  Recognition in this dataset often hinges on subtle color patterns, feather textures, and pose variations.

  \item \textbf{Flower102~\cite{nilsback2008automated}:} 
  Comprises 8,189 images from 102 flower species, posing challenges due to high inter-class visual similarity and variations in lighting, scale, and background clutter.
  It serves as a standard benchmark for evaluating fine-grained classification models in the botanical domain.

  \item \textbf{Food101~\cite{bossard2014food}:} 
  Consists of 101,000 images across 101 food categories, with each class containing 750 training and 250 test images. 
  The dataset exhibits large intra-class variability due to differences in cooking styles, presentation, and lighting, making it an excellent testbed for evaluating model robustness and generalization in the context of fine-grained food recognition.
\end{itemize}

\paragraph{Domain-Specific Tasks}
These datasets cover more specialized and domain-intensive classification tasks, such as texture analysis, medical diagnosis, and remote sensing, providing a rigorous test of a model's adaptability to non-standard visual domains:
\begin{itemize}
  \item \textbf{DTD~\cite{cimpoi2014describing}:} 
  Comprising 5,640 images grouped into 47 texture classes (e.g., striped, bubbly, porous), DTD focuses on recognizing mid-level perceptual patterns in textures. 
  Unlike traditional object-centric datasets, it evaluates models on their ability to capture material properties and fine-grained textural information, often independent of object identity.

  \item \textbf{HAM10000~\cite{tschandl2018ham10000}:} 
  A medically oriented dataset of 10,015 dermatoscopic images labeled into 7 diagnostic classes of skin lesions (e.g., melanoma, basal cell carcinoma). 
  HAM10000 is widely used in the medical imaging community for training and evaluating models in skin cancer classification. 
  It provides a realistic setting for assessing model reliability, especially in safety-critical applications.

  \item \textbf{Resisc45~\cite{cheng2017remote}:} 
  Composed of 31,500 satellite images across 45 scene categories such as airports, harbors, and residential areas, this dataset evaluates models' capabilities in scene understanding from aerial imagery. 
  The scenes vary widely in scale, orientation, and structural layout, making Resisc45 a comprehensive test for domain adaptation and robustness.
  
  \item \textbf{UCF101~\cite{soomro2012ucf101}:} 
  Originally designed for video-based human action recognition, UCF101 contains 13,320 videos spanning 101 action categories. 
  For our purposes, we extract representative still frames from these videos and repurpose the dataset for image-based action classification. 
  This approach allows us to evaluate the capacity of static models to infer dynamic actions, a challenging task due to the temporal nature of the source content.
\end{itemize}

%ps acc 270.3 cea 204.7
%dn acc 267.0 cea 177.1 
%vl acc 268.1 cea 200.1

\begin{table}[t]
\small
\centering
\resizebox{0.99\columnwidth}{!}{
\begin{tabular}{lrrrr}
  \toprule
  \textbf{Dataset} & \textbf{\# Classes} & \textbf{Train} & \textbf{Development} & \textbf{Test} \\
  \midrule
  Aircraft  & 100   & 3,334     & 3,333  & 3,333  \\
  CIFAR10   & 10    & 45,000    & 5,000  & 10,000 \\
  CIFAR100  & 100   & 45,000    & 5,000  & 10,000 \\
  CUB200    & 200   & 3,994     & 2,000  & 5,794  \\
  Flower102 & 102   & 4,093     & 1,633  & 2,463  \\
  Food101   & 101   & 50,500    & 20,200 & 30,300 \\
  DTD       & 47    & 2,820     & 1,128  & 1,692  \\
  HAM10000  & 7     & 8,010     & 1,000  & 1,005  \\
  ImageNet  & 1,000 & 1,281,167 & 50,000 & --     \\
  Resisc45  & 45    & 3,150     & 3,150  & 25,200 \\
  UCF101    & 101   & 7,639     & 1,898  & 3,783  \\
  \bottomrule
\end{tabular}}
\caption{Summary of dataset statistics and data splits used in our experiments, including the number of classes and the distribution across training, development, and test sets.}
\label{tab:dataset_splits}
\end{table}
\begin{table*}[t]
\small
\centering
\renewcommand{\arraystretch}{1.3}
\resizebox{0.99\textwidth}{!}{%
\begin{tabular}{llcccccccccc}
\toprule
\multirow{2}{*}{\textbf{Dataset}} 
& \multirow{2}{*}{\textbf{Backbone}} 
& \multirow{2}{*}{$K$}  
& \multirow{2}{*}{$\tau_{\text{merge}}$} 
& \multicolumn{4}{c}{\textbf{CBL}} 
& \multicolumn{3}{c}{\textbf{FCL}} \\
\cmidrule(lr){5-8} \cmidrule(lr){9-11}
& & & 
& \textbf{Batch Size} & \textbf{Max Steps} & \textbf{LR} & \textbf{Weight Decay} 
& \textbf{Batch Size} & \textbf{Max Iterations} & $\lambda$ \\
\midrule
\multirow{2}{*}{Aircraft}   & \texttt{CLIP\_RN50}       & 5 & 0.9997 & 32    & 20,000 & 5e-4    & 0       & 256 & 10,000 & 7e-4    \\
                            & \texttt{CLIP\_ViT-L/14}   & 4 & 0.9998 & 32    & 20,000 & 5e-4    & 0       & 256 & 10,000 & 7e-4    \\
\midrule
\multirow{2}{*}{CIFAR10}    &\texttt{CLIP\_RN50}        & 5 & 0.9996 & 1,024 & 20,000 & 4e-4    & 0       & 256 & 10,000 & 7e-4    \\
                            & \texttt{CLIP\_ViT-L/14}   & 5 & 0.9999 & 32    & 10,000 & 5e-4    & 0       & 256 & 10,000 & 7e-4    \\
\midrule
\multirow{2}{*}{CIFAR100}   & \texttt{CLIP\_RN50}       & 5 & 0.9997 & 512   & 20,000 & 5e-4    & 0       & 256 & 10,000 & 7e-4    \\
                            & \texttt{CLIP\_ViT-L/14}   & 5 & 0.9997 & 32    & 20,000 & 2e-4    & 0       & 256 & 10,000 & 7e-4    \\
\midrule
\multirow{2}{*}{CUB200}     & \texttt{CLIP\_RN50}       & 1 & 0.9998 & 32    & 30,000 & 2e-4    & 1e-5    & 256 & 10,000 & 8e-4    \\
                            & \texttt{CLIP\_ViT-L/14}   & 2 & 0.9998 & 32    & 30,000 & 1e-4    & 1e-5    & 256 & 10,000 & 9e-4    \\
\midrule
\multirow{2}{*}{DTD}        & \texttt{CLIP\_RN50}       & 1 & 0.9996 & 32    & 10,000 & 5e-4    & 0       & 256 & 10,000 & 7e-4    \\
                            & \texttt{CLIP\_ViT-L/14}   & 5 & 0.9996 & 32    &  5,000 & 1e-3    & 0       & 256 & 10,000 & 7e-4    \\
\midrule
\multirow{2}{*}{Flower102}  & \texttt{CLIP\_RN50}       & 1 & 0.9997 & 512   & 20,000 & 5e-4    & 0       & 256 & 10,000 & 7e-4    \\
                            & \texttt{CLIP\_ViT-L/14}   & 0 & 0.9996 & 32    & 30,000 & 1e-4    & 1e-5    & 256 & 10,000 & 2e-4    \\
\midrule
\multirow{2}{*}{Food101}    & \texttt{CLIP\_RN50}       & 5 & 0.9998 & 32    & 20,000 & 5e-4    & 0       & 256 & 10,000 & 7e-4    \\
                            & \texttt{CLIP\_ViT-L/14}   & 2 & 0.9996 & 32    & 20,000 & 1e-4    & 1e-5    & 256 & 10,000 & 7e-4    \\
\midrule
\multirow{2}{*}{HAM10000}   & \texttt{CLIP\_RN50}       & 3 & 0.9996 & 512   & 20,000 & 1e-3    & 0       & 256 & 10,000 & 7e-4    \\
                            & \texttt{CLIP\_ViT-L/14}   & 1 & 0.9996 & 32    & 20,000 & 5e-4    & 0       & 256 & 10,000 & 7e-4    \\
\midrule
\multirow{2}{*}{ImageNet}   & \texttt{CLIP\_RN50}       & 1 & 0.9997 & 1,024 &     60 & 1e-3    & 0       & 512 &    80 & 5e-4    \\
                            & \texttt{CLIP\_ViT-L/14}   & 1 & 0.9997 & 1,024 &     30 & 1e-4    & 0       & 512 &    80 & 1e-4    \\
\midrule
\multirow{2}{*}{Resisc45}   & \texttt{CLIP\_RN50}       & 5 & 0.9997 & 32    & 20,000 & 1e-3    & 0       & 256 & 10,000 & 7e-4    \\
                            & \texttt{CLIP\_ViT-L/14}   & 2 & 0.9998 & 32    &  5,000 & 4e-4    & 0       & 256 & 10,000 & 7e-4    \\
\midrule
\multirow{2}{*}{UCF101}     & \texttt{CLIP\_RN50}       & 5 & 0.9997 & 32    & 20,000 & 5e-4    & 0       & 256 & 10,000 & 7e-4    \\
                            & \texttt{CLIP\_ViT-L/14}   & 2 & 0.9997 & 32    & 20,000 & 1e-4    & 1e-5    & 256 & 10,000 & 7e-4    \\
\bottomrule
\end{tabular}}
\caption{Hyperparameter settings for the Concept Bottleneck Layer (CBL) and Final Classification Layer (FCL) in \textbf{PS-CBM} across all datasets, using \texttt{CLIP\_RN50} and \texttt{CLIP\_ViT-L/14} backbones. Note: LR is short for Learning Rate; for ImageNet, Max Steps and Max Iterations indicate the number of epochs.}
\label{tab:param_ps-cbm}
\end{table*}

\begin{table*}[t]
\centering
\small
\setlength\tabcolsep{3pt}
\renewcommand{\arraystretch}{1.5}
\resizebox{0.99\textwidth}{!}{%
\begin{tabular}{l *{11}{cc}}
\toprule
\multirow{2.5}{*}{\textbf{Method}} 
& \multicolumn{2}{c}{\textbf{Aircraft}}
& \multicolumn{2}{c}{\textbf{CIFAR10}}
& \multicolumn{2}{c}{\textbf{CIFAR100}}
& \multicolumn{2}{c}{\textbf{CUB200}}
& \multicolumn{2}{c}{\textbf{DTD}}
& \multicolumn{2}{c}{\textbf{Flower102}}
& \multicolumn{2}{c}{\textbf{Food101}}
& \multicolumn{2}{c}{\textbf{HAM10000}}
& \multicolumn{2}{c}{\textbf{ImageNet}}
& \multicolumn{2}{c}{\textbf{Resisc45}}
& \multicolumn{2}{c}{\textbf{UCF101}} \\
\cmidrule(lr){2-3} \cmidrule(lr){4-5} \cmidrule(lr){6-7} \cmidrule(lr){8-9} \cmidrule(lr){10-11} 
\cmidrule(lr){12-13} \cmidrule(lr){14-15} \cmidrule(lr){16-17} \cmidrule(lr){18-19} \cmidrule(lr){20-21} \cmidrule(lr){22-23}
& \textbf{ACC $\uparrow$} & \textbf{CEA $\uparrow$} 
& \textbf{ACC $\uparrow$} & \textbf{CEA $\uparrow$} 
& \textbf{ACC $\uparrow$} & \textbf{CEA $\uparrow$} 
& \textbf{ACC $\uparrow$} & \textbf{CEA $\uparrow$} 
& \textbf{ACC $\uparrow$} & \textbf{CEA $\uparrow$} 
& \textbf{ACC $\uparrow$} & \textbf{CEA $\uparrow$} 
& \textbf{ACC $\uparrow$} & \textbf{CEA $\uparrow$} 
& \textbf{ACC $\uparrow$} & \textbf{CEA $\uparrow$} 
& \textbf{ACC $\uparrow$} & \textbf{CEA $\uparrow$} 
& \textbf{ACC $\uparrow$} & \textbf{CEA $\uparrow$} 
& \textbf{ACC $\uparrow$} & \textbf{CEA $\uparrow$} \\
\midrule
\textbf{Linear Probe} 
& 62.9 & -- & 98.1 & -- & 87.1 & -- & 84.5 & -- & 81.1 & -- & 99.4 & -- 
& 93.2 & -- & 81.5 & -- & 84.3 & -- & 93.3 & -- & 89.8 & -- \\
\midrule
\textbf{LaBo} 
& 61.4 & 42.4 & 97.7 & 67.1 & 85.9 & 59.4 & 82.0 & 56.5 & 77.0 & 53.4 & 99.3 & 68.6 
& 92.4 & 63.9 & 80.5 & 53.0 & 84.1 & 57.1 & 91.5 & 63.5 & 90.2 & 62.3 \\
\textbf{LF-CBM} 
& 53.2 & 42.3 & 97.1 & 69.9 & 86.1 & 62.3 & 77.4 & 60.2 & 76.6 & 59.0 & 98.1 & 75.8 
& 91.8 & 69.5 & 72.3 & 58.1 & 79.8 & 57.7 & 91.8 & 67.6 & 88.4 & 64.3 \\
\textbf{LM4CV} 
& 54.7 & 40.9 & 96.7 & \underline{74.6} & 84.5 & 63.2 & 79.3 & 58.7 & 79.4 & \underline{60.1} & 98.8 & 73.8 
& 92.2 & 68.9 & 66.8 & 49.8 & \textbf{84.7} & 61.1 & 92.3 & 70.0 & 89.3 & 66.8 \\
\textbf{DN-CBM} 
& 62.9 & 43.2 & 97.7 & 61.7 & 86.3 & 59.3 & 83.3 & 58.2 & \underline{81.6} & 54.9 & \textbf{99.5} & 68.4 
& 92.2 & 63.4 & 81.2 & 48.4 & 79.1 & 56.7 & \underline{93.4} & 62.9 & 89.8 & 61.7 \\
\textbf{Res-CBM} 
& 54.4 & 42.7 & 97.5 & 68.9 & 83.0 & 62.7 & 75.8 & 60.1 & 73.6 & 56.5 & 98.6 & \underline{76.9} 
& 89.8 & \underline{69.8} & 76.8 & 52.2 & 80.6 & \underline{61.6} & 91.1 & \underline{68.5} & 88.0 & \underline{68.5} \\
\textbf{VLG-CBM} 
& \underline{63.5} & \underline{48.3} & 97.9 & 73.6 & \underline{87.0} & \underline{64.7} & \underline{84.5} & \underline{63.4} & 77.4 & 59.3 & 99.5 & 74.2 
& \underline{92.8} & 68.6 & \underline{82.0} & \underline{61.5} & 82.5 & 59.8 & 92.4 & 70.8 & \underline{90.2} & 67.4 \\
\textbf{V2C-CBM} 
& 57.9 & 40.0 & \underline{98.0} & 67.3 & 86.1 & 59.5 & 80.8 & 55.7 & 79.8 & 55.3 & 99.2 & 70.0 
& 92.2 & 63.7 & 78.9 & 51.9 & 84.3 & 57.3 & 92.6 & 64.3 & 88.2 & 61.0 \\
\textbf{DCBM} 
& 57.9 & 41.2 & 97.5 & 63.7 & 85.2 & 60.6 & 80.8 & 58.4 & 78.4 & 54.6 & 99.1 & 70.4 
& 92.0 & 65.4 & 78.0 & 48.1 & 76.5 & 56.7 & 91.4 & 63.6 & 87.5 & 62.2 \\
\textbf{Ours} 
& \textbf{65.1} & \textbf{48.5} & \textbf{98.0} & \textbf{74.8} & \textbf{87.3} & \textbf{64.9} & \textbf{85.3} & \textbf{64.1} & \textbf{82.1} & \textbf{61.5} & \underline{99.5} & \textbf{77.7} 
& \textbf{93.0} & \textbf{71.1} & \textbf{82.8} & \textbf{63.9} & \underline{84.5} & \textbf{62.3} & \textbf{93.4} & \textbf{72.4} & \textbf{90.4} & \textbf{69.7} \\
\bottomrule
\end{tabular}}
\caption{ACC and CEA on 11 datasets using \texttt{CLIP\_ViT-L/14} backbone. \textbf{Bold} and \underline{underline} denote the best and second-best results, respectively.}
\label{tab:acc_cea_results_vit}
\end{table*}
\begin{table}[t]
\small
\centering
\renewcommand{\arraystretch}{1.3}
\resizebox{0.99\linewidth}{!}{%
\begin{tabular}{lrrr}
  \toprule
  \textbf{Method} & \textbf{Avg. ACC (\%) $\uparrow$} & \textbf{Avg. \# Concepts $\downarrow$} & \textbf{Avg. CEA (\%) $\uparrow$} \\
  \midrule
  \textbf{LaBo}    & 85.6             & 8,241          & 58.8 \\
  \textbf{LF-CBM}  & 83.0             & 787            & 62.4 \\
  \textbf{LM4CV}   & 83.5             & 824            & 62.5 \\
  \textbf{DN-CBM}  & 86.1             & 6,144          & 58.1 \\
  \textbf{Res-CBM} & 82.7             & \textbf{282}   & 62.6 \\
  \textbf{VLG-CBM} & \underline{86.4} & 739            & \underline{64.7} \\
  \textbf{V2C-CBM} & 85.3             & 8,009          & 58.7 \\
  \textbf{DCBM}    & 84.0             & 2,048          & 58.6 \\
  \textbf{PS-CBM}  & \textbf{87.4}    & \underline{497}   & \textbf{66.4} \\
  \bottomrule
\end{tabular}}
\caption{Comparison of methods by average ACC, number of concepts used, and CEA using \texttt{CLIP\_ViT-L/14} backbone. \textbf{Bold} and \underline{underlined} indicate the best and second-best results, respectively.}
\label{tab:average_acc_cea_vit}
\end{table}

\begin{table}[t]
\centering
\small
\setlength\tabcolsep{3pt}
\renewcommand{\arraystretch}{1.3}
\begin{tabular}{l *{3}{cc}}
\toprule
\multirow{2.5}{*}{\textbf{Method}} 
& \multicolumn{2}{c}{\textbf{CIFAR10}}
& \multicolumn{2}{c}{\textbf{Flower102}}
& \multicolumn{2}{c}{\textbf{Food101}} \\
\cmidrule(lr){2-3} \cmidrule(lr){4-5} \cmidrule(lr){6-7}
& \textbf{ACC $\uparrow$} & \textbf{CEA $\uparrow$} 
& \textbf{ACC $\uparrow$} & \textbf{CEA $\uparrow$} 
& \textbf{ACC $\uparrow$} & \textbf{CEA $\uparrow$} \\
\midrule
\textbf{DN-CBM} 
& 88.2 & 55.2 & 96.5 & 65.8 & \underline{82.3} & 56.1 \\
\textbf{VLG-CBM} 
& \underline{89.5} & \underline{67.4} & \underline{97.1} & \underline{72.4} & 81.5 & \underline{60.3} \\
\textbf{PS-CBM} 
& \textbf{89.7} & \textbf{68.4} & \textbf{97.8} & \textbf{74.8} & \textbf{82.8} & \textbf{61.5} \\
\bottomrule
\end{tabular}
\caption{Error bar analysis on the \texttt{CLIP\_RN50} backbone using five random seeds (std $<$ 0.2\%).}
\label{tab:acc_cea_error_bars}
\end{table}
\begin{table*}[ht]
\small
\centering
\setlength\tabcolsep{3pt} % 平衡列间距
\renewcommand{\arraystretch}{1.2}
\resizebox{\linewidth}{!}{%
\begin{tabular}{lll c*{11}{c}}
\toprule
\multirow{3}{*}{\textbf{Method}} & \multirow{3}{*}{\textbf{Backbone}} & \multirow{3}{*}{\textbf{Param}} & \multicolumn{11}{c}{\textbf{Datasets}} \\
\cmidrule(lr){4-14}
& & & \textbf{Aircraft} & \textbf{CIFAR10} & \textbf{CIFAR100} & \textbf{CUB200} & \textbf{DTD} & \textbf{Flower102} & \textbf{Food101} & \textbf{HAM10000} & \textbf{ImageNet} & \textbf{Resisc45} & \textbf{UCF101} \\
\midrule
\multirow{6}{*}{LaBo} 
& \multirow{3}{*}{\texttt{CLIP\_RN50}} 
  & BS    & 256 & 512 & 512 & 512 & 512 & 256 & 1,024 & 256 & 2,048 & 256 & 256 \\
& 
  & LR & 5e-5 & 1e-4 & 1e-4 & 5e-5 & 1e-4 & 1e-5 & 1e-5 & 5e-4 & 1e-5 & 5e-5 & 1e-5 \\
& 
  & E/S & 50,000 & 15,000 & 50,000 & 15,000 & 15,000 & 50,000 & 50,000 & 20,000 & 3,000 & 15,000 & 50,000 \\
\cmidrule(lr){2-14}
& \multirow{3}{*}{\texttt{CLIP\_ViT-L/14}} 
  & BS    & 256 & 512 & 512 & 512 & 512 & 256 & 1,024 & 256 & 2,048 & 256 & 256 \\
& 
  & LR & 5e-5 & 1e-4 & 1e-5 & 5e-5 & 1e-4 & 1e-5 & 1e-5 & 5e-4 & 1e-5 & 5e-5 & 1e-5 \\
& 
  & E/S & 10,000 & 15,000 & 10,000 & 5,000 & 15,000 & 50,000 & 5,000 & 10,000 & 3,000 & 15,000 & 50,000 \\
\midrule
\multirow{6}{*}{LF-CBM} 
& \multirow{3}{*}{\texttt{CLIP\_RN50}} 
  & BS    & 256 & 256 & 256 & 256 & 256 & 256 & 256 & 256 & 256 & 256 & 256 \\
& 
  & LR & 1e-3 & 1e-3 & 1e-3 & 1e-3 & 1e-3 & 1e-3 & 1e-3 & 1e-3 & 1e-3 & 1e-3 & 1e-3 \\
& 
  & E/S & 5,000 & 5,000 & 5,000 & 5,000 & 5,000 & 5,000 & 5,000 & 5,000 & 5,000 & 5,000 & 5,000 \\
\cmidrule(lr){2-14}
& \multirow{3}{*}{\texttt{CLIP\_ViT-L/14}} 
  & BS    & 256 & 256 & 256 & 256 & 256 & 256 & 256 & 256 & 256 & 256 & 256 \\
& 
  & LR & 1e-3 & 1e-3 & 1e-3 & 1e-3 & 1e-3 & 1e-3 & 1e-3 & 1e-3 & 1e-3 & 1e-3 & 1e-3 \\
& 
  & E/S & 5,000 & 5,000 & 5,000 & 5,000 & 5,000 & 5,000 & 5,000 & 5,000 & 5,000 & 5,000 & 5,000 \\
\midrule
\multirow{6}{*}{LM4CV} 
& \multirow{3}{*}{\texttt{CLIP\_RN50}} 
  & BS    & 4,096 & 4,096 & 4,096 & 4,096 & 4,096 & 4,096 & 4,096 & 4,096 & 4,096 & 4,096 & 4,096 \\
& 
  & LR & 1e-2 & 1e-2 & 1e-2 & 1e-2 & 1e-2 & 1e-2 & 1e-2 & 1e-2 & 1e-2 & 1e-2 & 1e-2 \\
& 
  & E/S & 3,000 & 3,000 & 3,000 & 3,000 & 3,000 & 3,000 & 3,000 & 3,000 & 3,000 & 3,000 & 3,000 \\
\cmidrule(lr){2-14}
& \multirow{3}{*}{\texttt{CLIP\_ViT-L/14}} 
  & BS    & 4,096 & 4,096 & 4,096 & 4,096 & 4,096 & 4,096 & 4,096 & 4,096 & 4,096 & 4,096 & 4,096 \\
& 
  & LR & 1e-2 & 1e-2 & 1e-2 & 1e-2 & 1e-2 & 1e-2 & 1e-2 & 1e-2 & 1e-2 & 1e-2 & 1e-2 \\
& 
  & E/S & 3,000 & 3,000 & 3,000 & 3,000 & 3,000 & 3,000 & 3,000 & 3,000 & 3,000 & 3,000 & 3,000 \\
\midrule
\multirow{6}{*}{DN-CBM} 
& \multirow{3}{*}{\texttt{CLIP\_RN50}} 
  & BS    & 512 & 512 & 512 & 512 & 512 & 512 & 512 & 512 & 512 & 512 & 512 \\
& 
  & LR & 1e-2 & 1e-2 & 1e-2 & 1e-2 & 1e-2 & 1e-2 & 1e-2 & 1e-2 & 1e-2 & 1e-2 & 1e-2 \\
& 
  & E/S & 500 & 500 & 500 & 500 & 500 & 500 & 500 & 500 & 1,000 & 500 & 500 \\
\cmidrule(lr){2-14}
& \multirow{3}{*}{\texttt{CLIP\_ViT-L/14}} 
  & BS    & 512 & 512 & 512 & 512 & 512 & 512 & 512 & 512 & 512 & 512 & 512 \\
& 
  & LR & 1e-2 & 1e-2 & 1e-2 & 1e-2 & 1e-2 & 1e-2 & 1e-2 & 1e-2 & 1e-2 & 1e-2 & 1e-2 \\
& 
  & E/S & 500 & 500 & 500 & 500 & 500 & 500 & 500 & 500 & 1,000 & 500 & 500 \\
\midrule
\multirow{6}{*}{Res-CBM} 
& \multirow{3}{*}{\texttt{CLIP\_RN50}} 
  & BS    & 128 & 128 & 128 & 128 & 128 & 128 & 128 & 128 & 1,024 & 128 & 128 \\
& 
  & LR & 1e-2 & 1e-2 & 1e-2 & 1e-2 & 1e-2 & 1e-2 & 1e-2 & 1e-2 & 1e-2 & 1e-2 & 1e-2 \\
& 
  & E/S & 50 & 50 & 50 & 50 & 50 & 50 & 50 & 50 & 50 & 50 & 50 \\
\cmidrule(lr){2-14}
& \multirow{3}{*}{\texttt{CLIP\_ViT-L/14}} 
  & BS    & 128 & 128 & 128 & 128 & 128 & 128 & 128 & 128 & 1,024 & 128 & 128 \\
& 
  & LR & 1e-2 & 1e-2 & 1e-2 & 1e-2 & 1e-2 & 1e-2 & 1e-2 & 1e-2 & 1e-2 & 1e-2 & 1e-2 \\
& 
  & E/S & 50 & 50 & 50 & 50 & 50 & 50 & 50 & 50 & 50 & 50 & 50 \\
\midrule
\multirow{6}{*}{VLG-CBM} 
& \multirow{3}{*}{\texttt{CLIP\_RN50}} 
  & BS    & 32 & 32 & 32 & 32 & 32 & 32 & 32 & 32 & 64 & 32 & 32 \\
& 
  & LR & 5e-4 & 5e-4 & 5e-4 & 5e-4 & 5e-4 & 5e-4 & 5e-4 & 5e-4 & 1e-2 & 5e-4 & 5e-4 \\
& 
  & E/S & 300 & 100 & 100 & 100 & 100 & 300 & 200 & 100 & 20 & 100 & 100 \\
\cmidrule(lr){2-14}
& \multirow{3}{*}{\texttt{CLIP\_ViT-L/14}} 
  & BS    & 32 & 32 & 32 & 32 & 32 & 32 & 32 & 32 & 64 & 32 & 32 \\
& 
  & LR & 5e-4 & 5e-4 & 5e-4 & 5e-4 & 5e-4 & 5e-4 & 5e-4 & 5e-4 & 1e-2 & 5e-4 & 5e-4 \\
& 
  & E/S & 100 & 10 & 10 & 100 & 20 & 100 & 15 & 100 & 15 & 100 & 100 \\
\midrule
\multirow{6}{*}{V2C-CBM} 
& \multirow{3}{*}{\texttt{CLIP\_RN50}} 
  & BS    & 256 & 512 & 512 & 512 & 512 & 256 & 1,024 & 256 & 1,024 & 256 & 256 \\
& 
  & LR & 5e-6 & 1e-4 & 1e-5 & 5e-5 & 1e-4 & 5e-5 & 1e-5 & 5e-4 & 1e-5 & 5e-5 & 1e-5 \\
& 
  & E/S & 10,000 & 50,000 & 50,000 & 15,000 & 15,000 & 20,000 & 50,000 & 20,000 & 2,000 & 15,000 & 50,000 \\
\cmidrule(lr){2-14}
& \multirow{3}{*}{\texttt{CLIP\_ViT-L/14}} 
  & BS    & 256 & 512 & 512 & 512 & 512 & 256 & 1,024 & 256 & 1,024 & 256 & 256 \\
& 
  & LR & 5e-5 & 1e-4 & 1e-5 & 5e-5 & 1e-4 & 1e-5 & 1e-5 & 5e-4 & 1e-5 & 5e-5 & 1e-5 \\
& 
  & E/S & 10,000 & 50,000 & 50,000 & 15,000 & 15,000 & 20,000 & 50,000 & 20,000 & 2,000 & 15,000 & 50,000 \\
\midrule
\multirow{6}{*}{DCBM} 
& \multirow{3}{*}{\texttt{CLIP\_RN50}} 
  & BS    & 128 & 128 & 128 & 128 & 128 & 128 & 128 & 128 & 256 & 128 & 128 \\
& 
  & LR & 1e-4 & 1e-4 & 1e-4 & 1e-4 & 1e-4 & 1e-4 & 1e-4 & 1e-4 & 1e-4 & 1e-4 & 1e-4 \\
& 
  & E/S & 1,000 & 1,000 & 1,000 & 1,000 & 1,000 & 1,000 & 1,000 & 1,000 & 1,000 & 1,000 & 1,000 \\
\cmidrule(lr){2-14}
& \multirow{3}{*}{\texttt{CLIP\_ViT-L/14}} 
  & BS    & 128 & 128 & 128 & 128 & 128 & 128 & 128 & 128 & 256 & 128 & 128 \\
& 
  & LR & 1e-4 & 1e-4 & 1e-4 & 1e-4 & 1e-4 & 1e-4 & 1e-4 & 1e-4 & 1e-4 & 1e-4 & 1e-4 \\
& 
  & E/S & 1,000 & 1,000 & 1,000 & 1,000 & 1,000 & 1,000 & 1,000 & 1,000 & 1,000 & 1,000 & 1,000 \\
\bottomrule
\end{tabular}%
}
\caption{Hyperparameter settings for all baselines across all 11 datasets, including Batch Size (BS), Learning Rate (LR), and Epochs/ Steps  (E/S). Each method is evaluated with two backbones (\texttt{CLIP\_RN50} and \texttt{CLIP\_ViT-L/14}), and each backbone is represented by three rows corresponding to different training components.}
\label{tab:param_baslines}
\end{table*}

\subsection{Dataset Splits}
\label{appendix:impl:splits}

To ensure fair and consistent evaluation across all baselines and experimental settings, we adopt standard or widely accepted dataset splits, summarized in Table~\ref{tab:dataset_splits}. 
%%%
The splitting strategies are as follows:
\begin{itemize}
  \item For \textbf{Aircraft}, \textbf{DTD}, \textbf{Flower102}, \textbf{Food101}, \textbf{Resisc45}, and \textbf{UCF101}, we use the official splits provided by LaBo~\cite{yang2023language}, which balance class representation and enable reproducibility.

  \item For \textbf{CUB200}, we randomly sample 10 training images per class to construct a development set, maintaining class balance with a minimal validation budget.

  \item For \textbf{CIFAR10} and \textbf{CIFAR100}, we randomly reserve 10\% of the training data for development, following standard practice in few-shot and semi-supervised settings.

  \item For \textbf{HAM10000}, we apply a stratified 80/10/10 split per class for training, development, and test, preserving diagnostic class balance.

  \item For \textbf{ImageNet}, we evaluate only on the development set due to the scale and computational cost of full training.
\end{itemize}

\begin{figure}[t]
  \centering
  \includegraphics[width=\columnwidth]{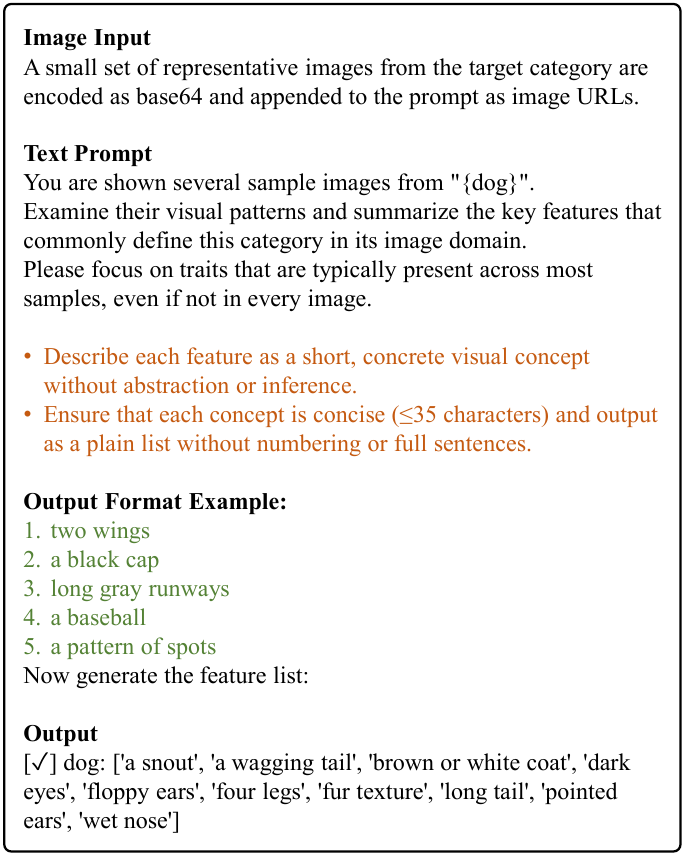}
  \caption{Example of the multimodal concept generation prompt for the class ``dog'' in the CIFAR10 dataset. The prompt combines representative images with a natural language instruction to elicit visually grounded, semantically meaningful concepts from GPT.}
  \label{fig:prompt}
\end{figure}

\subsection{Prompt Design}
\label{appendix:impl:prompt}

To generate visually grounded concepts for each category, we design a unified prompting strategy that integrates concise textual instructions with exemplar images. 
The textual component is crafted to elicit descriptions of distinctive and observable visual features, deliberately avoiding abstract reasoning or domain-specific jargon. 

For each category, we interact with GPT using \emph{four} representative images and issue \emph{two} separate queries. 
The resulting concept sets are then \emph{merged and deduplicated} to produce a concise and diverse list of relevant concepts. 
This dual-query setup helps capture a broader range of semantic attributes while ensuring consistency across responses.

Figure~\ref{fig:prompt} provides an illustrative example of the prompt used for the class ``dog'' in CIFAR10, highlighting the structure and effectiveness of multimodal concept generation.

\subsection{Hyperparameter Settings}
\label{appendix:impl:params_ps-cbm}

\paragraph{Parameter Settings for PS-CBM}
In the experiments, we set $\tau_{\text{conf}} = 0.20$ for concept filtering and labeling and $\beta=0.25$ when computing the CEA. For the rest hyperparameters, we summarize the settings in Table~\ref{tab:param_ps-cbm}.

\paragraph{Parameter Settings for Baseline Methods}
Table~\ref{tab:param_baslines} summarizes the hyperparameter configurations used for all baseline methods. 
Below, we detail the specific settings for each:
\begin{itemize}
  \item \textbf{LaBo~\cite{yang2023language}:} 
  We use the concept set provided by the original authors and enable the \texttt{remove\_cls\_name} option to exclude class names from candidate concepts. 
  For CIFAR100, the learning rate is increased from 1e-5 to 1e-4. All datasets are trained for extended epochs to ensure convergence, while other hyperparameters remain at their defaults.

  \item \textbf{LF-CBM~\cite{oikarinen2023label}:} 
  Concepts are generated using \texttt{GPT-3.5-turbo-instruct}. We tune key hyperparameters such as \texttt{lam}, \texttt{clip\_cutoff}, and \texttt{interpretability\_cutoff} via grid search. Other training configurations are standardized to ensure efficiency and consistency across datasets.

  \item \textbf{LM4CV~\cite{yan2023learning}:} 
  Concepts are generated using \texttt{GPT-3.5-turbo-instruct}. We set the number of attributes to five times the number of classes for each dataset. 
  The learning rate is fixed at 0.01, batch size at 4,096, and both training and linear probing epochs at 3,000, with early stopping enabled. 
  Mahalanobis distance regularization is activated with strength 1.

  \item \textbf{DN-CBM~\cite{rao2024discover}:} 
  Concepts are drawn from a 20,000-word vocabulary adopted by CLIP-Dissect~\cite{oikarinen2022clip}, publicly available online.\footnote{\url{https://github.com/first20hours/google-10000-english/blob/master/20k.txt}}. 
  We adjust the number of probe epochs and learning rate to balance training efficiency and quality. All other settings follow default values.

  \item \textbf{Res-CBM~\cite{shang2024incremental}:} 
  Concepts come from dictionaries, which are built from a base concept bank combining standard TCAV concepts and ConceptNet-derived associations, together with a larger candidate pool extracted from Visual Genome. 
  The residual dimension is set to 30 for ImageNet and 15 for other datasets. Following the original paper, similarity regularization is set to 1 and the candidate number to 5. 
  Additional tuning ensures timely convergence.

  \item \textbf{VLG-CBM~\cite{srivastava2024vlg}:} 
  Concepts are generated by \texttt{GPT-3.5-turbo-instruct}, with ImageNet annotations provided by the authors. 
  The number of hidden layers in the concept bottleneck layer is fixed to one to improve expressiveness and accuracy. 
  The regularization parameter \texttt{saga lam} is set to 0.0007 (or 0.0005 for ImageNet). 
  Automatic weighting is enabled with a positive weight of 0.2 to address class imbalance. 
  The probability of cropping to concept varies: 0.3--0.5 for fine-grained datasets (Aircraft, Flower, Food, HAM10000, and CUB), and 0 otherwise.
  Additional parameters such as learning rate, epochs, and batch size are tuned per dataset.

  \item \textbf{V2C-CBM~\cite{he2025v2c}:} 
  Concepts (500 per class) are provided by the authors. Weight matrices are randomly initialized, and each class uses 50 concepts uniformly. All other settings follow the original implementation.

  \item \textbf{DCBM~\cite{prasse2025dcbm}:}
  This method uses the same 20,000-word vocabulary as DN-CBM and employs Mask R-CNN for concept generation. Concepts are clustered into 2,048 groups using k-means with median-based centroids. For ImageNet, the batch size is increased to 256 to accommodate the dataset scale.
\end{itemize}

\subsection{Error Bar Analysis on \texttt{CLIP\_RN50} Backbone}
\label{appendix:rn50_error_bars}

To evaluate the robustness and statistical significance of the main results, 
we conduct additional experiments using five random seeds on three datasets with the smallest performance margins—CIFAR10, Flower102, and Food101—under the \texttt{CLIP\_RN50} backbone.
As shown in Table~\ref{tab:acc_cea_error_bars}, PS-CBM consistently surpasses competing CBMs with minimal variance across runs.
Specifically, PS-CBM still outperforms DN-CBM by +1.12\% in ACC and +9.22\% in CEA, and exceeds VLG-CBM by +0.73\% in ACC and +1.60\% in CEA, 
with all standard deviations below 0.2\%.
These results confirm the stability and statistical reliability of PS-CBM’s performance improvements.

\section{Results with \texttt{CLIP\_ViT-L/14} Backbone}
\label{appendix:vit_results}

To validate the robustness of our findings across architectures, we replicate the main experiments using the more powerful \texttt{CLIP\_ViT-L/14} backbone, in addition to the default \texttt{CLIP\_RN50}.
%%%
The performance of PS-CBM and all baselines is summarized in Tables \ref{tab:acc_cea_results_vit} and \ref{tab:average_acc_cea_vit}.

Consistent with the RN50-based results, PS-CBM continues to demonstrate strong generalization and concept efficiency.
%%%
As shown in Table \ref{tab:average_acc_cea_vit}, on average across 11 datasets, it outperforms state-of-the-art CBMs by 1.0\%--4.7\% in classification accuracy (ACC) and by 1.7\%--8.3\% in Concept-Efficient Accuracy (CEA).
%%%
From Table \ref{tab:acc_cea_results_vit}, at the dataset level, PS-CBM consistently achieves the highest CEA across all benchmarks, underscoring its ability to deliver interpretable predictions without compromising performance.
It also secures the top ACC score on most datasets, with only minor deviations on Flower102 and ImageNet, where it ranks second with marginal gaps.

Importantly, PS-CBM accomplishes these results while using substantially fewer concepts compared to methods like LM4CV and DN-CBM, which rely on much larger concept sets to reach similar accuracy.   
This demonstrates that PS-CBM not only maintains competitive predictive performance but also ensures greater interpretability and compactness, reducing redundancy and mitigating the risk of concept leakage.

%\newpage
%\input{09_rebuttal_simplify}

\end{document}